\pdfoutput=1
\documentclass[11pt]{article}

\usepackage[final]{acl}

\usepackage{times}
\usepackage{latexsym}

\usepackage[T1]{fontenc}
\usepackage[utf8]{inputenc}

\usepackage{microtype}

\usepackage{inconsolata}

\usepackage{graphicx}

\usepackage{hyperref}
\usepackage{url}
\usepackage{subcaption}
\usepackage{amsfonts}
\usepackage[utf8]{inputenc} % allow utf-8 input
\usepackage[T1]{fontenc}    % use 8-bit T1 fonts
\usepackage{booktabs}
\usepackage{graphicx}
\usepackage{multirow}
\usepackage{multicol}
\usepackage{enumitem}
\usepackage{amsmath}
\usepackage{microtype}
\usepackage{caption}
\usepackage{wrapfig}
\usepackage{algorithm}
\usepackage{algpseudocode}
\usepackage{listings}
\usepackage{xcolor,colortbl}
\usepackage{csquotes}
\usepackage{pifont}
\usepackage{arydshln}
\usepackage{framed}

\usepackage{etoolbox}
\AtBeginEnvironment{multline}{
\setlength{\abovedisplayskip}{2pt}
\setlength{\belowdisplayskip}{2pt}
\setlength{\abovedisplayshortskip}{1pt}
\setlength{\belowdisplayshortskip}{1pt}
}

\newcommand{\cmark}{\ding{51}}%
\newcommand{\xmark}{\ding{55}}%

\definecolor{b}{RGB}{70,130,180}
\definecolor{r}{RGB}{220,20,60}
\definecolor{g}{RGB}{34,139,34}
\definecolor[named]{ACMPurple}{cmyk}{0.55,1,0,0.15}
\definecolor[named]{ACMDarkBlue}{cmyk}{1,0.58,0,0.21}
\newcommand{\ours}{OKGQA}
\newcommand{\oursp}{OKGQA-P}
\newcommand{\codelink}{\url{https://github.com/Y-Sui/OKGQA}}

\newcommand{\highlight}[1]{\textcolor{brown}{\textbf{#1}}}
\newcommand{\highlightEx}[1]{\textcolor{teal}{\textbf{#1}}}

\newcommand{\refsec}[1]{\S\ref{#1}} % \textsection

\title{Can Knowledge Graphs Make Large Language Models More Trustworthy? An Empirical Study Over Open-ended Question Answering}

\makeatletter
\renewcommand*{\@fnsymbol}[1]{\ensuremath{\ifcase#1\or *\or \dagger\or \ddagger\or
   \mathsection\or \mathparagraph\or \|\or **\or \dagger\dagger
   \or \ddagger\ddagger \else\@ctrerr\fi}}
\makeatother

\author{
Yuan Sui\textsuperscript{\rm 1}, 
Yufei He\textsuperscript{\rm 1}, 
Zifeng Ding\textsuperscript{\rm 2}, 
Bryan Hooi\textsuperscript{\rm 1}\\
\textsuperscript{\rm 1} National University of Singapore
\textsuperscript{\rm 2} University of Cambridge\\
\texttt{\{\href{mailto:yuansui@comp.nus.edu.sg}{yuansui}, \href{mailto:yufei.he@comp.nus.edu.sg}{yufei.he}, \href{mailto:bhooi@comp.nus.edu.sg}{bhooi}\}@comp.nus.edu.sg},
\texttt{\href{mailto:zd320@cam.ac.uk}{zd320@cam.ac.uk}}}

\begin{document}
\maketitle

\begin{abstract}

Recent works integrating Knowledge Graphs (KGs) have shown promising improvements in enhancing the reasoning capabilities of Large Language Models (LLMs). However, existing benchmarks primarily focus on closed-ended tasks, leaving a gap in evaluating performance on more complex, real-world scenarios. This limitation also hinders a thorough assessment of KGs' potential to reduce hallucinations in LLMs. To address this, we introduce OKGQA\footnote{Code and data are released at \codelink{}}, a new benchmark specifically designed to evaluate LLMs augmented with KGs in open-ended, real-world question answering settings. 
OKGQA reflects practical complexities through diverse question types and incorporates metrics to quantify both hallucination rates and reasoning improvements in LLM+KG models.
% OKGQA is designed to closely reflect the complexities of practical applications by utilizing diverse question from different types and incorporates specific metrics to measure both hallucination ratio and the enhancement in reasoning capabilities. 
To consider the scenarios in which KGs may contain varying levels of errors, we propose a benchmark variant, OKGQA-P, to assess model performance when the semantics and structure of KGs are deliberately perturbed and contaminated. In this paper, we aims to (1) explore whether KGs can make LLMs more trustworthy in an open-ended setting, and (2) conduct a comparative analysis to shed light on method design. We believe this study can facilitate a more complete performance comparison and encourages continuous improvement in integrating KGs with LLMs to mitigate hallucination, and make LLMs more trustworthy.

\end{abstract}

\section{Introduction}
\label{sec:introduction}

Contemporary large language models (LLMs) are prone to producing \textbf{hallucinations}—plausible-sounding but incorrect or irrelevant outputs—due to gaps and inconsistencies in their training data~\cite{gekhman2024doesfinetuningllmsnew, lee2023factualityenhancedlanguagemodels}. These inaccuracies often arise from misinformation, biases, or errors embedded in the data~\cite{weng2024hallucination,chen2025can}, posing significant risks in high-stakes domains such as healthcare~\cite{he2023surveyhealthcare} and scientific research~\cite{taylor2022galactica}.

\begin{figure}
    \centering
    \includegraphics[width=\linewidth]{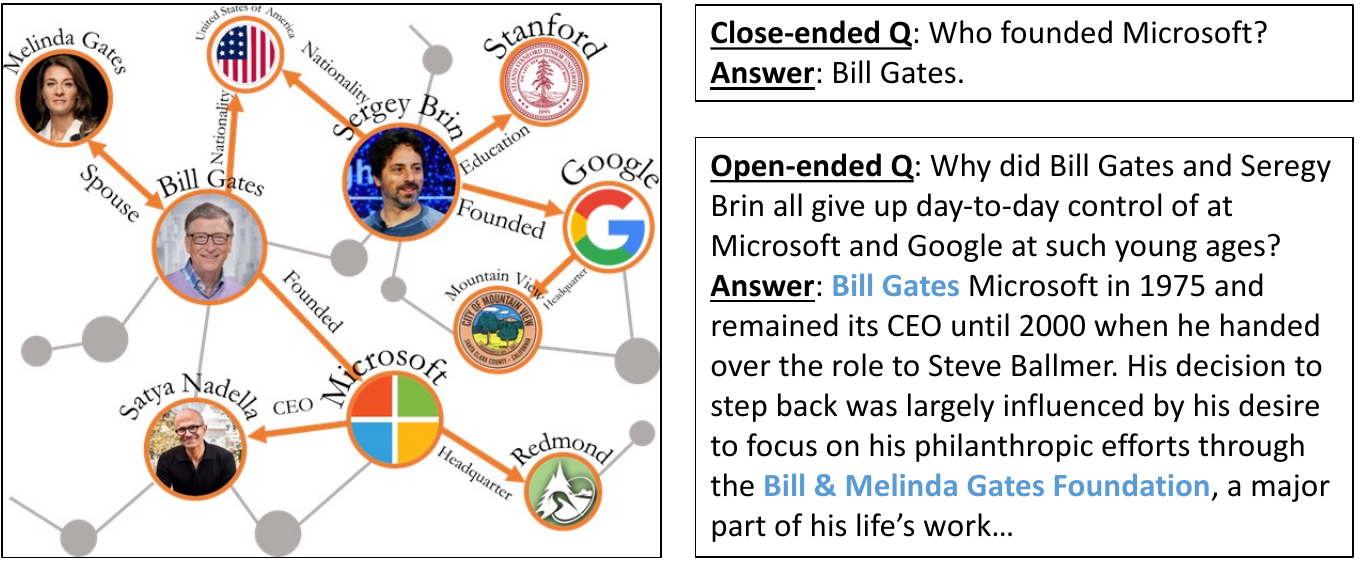}
    \caption{\small Comparison between open-ended and closed-ended questions over knowledge graphs.}
    \label{fig:demo}
    \vspace{-2mm}
\end{figure}

To mitigate these limitations, recent research has explored augmenting LLMs with external knowledge graphs (KGs)~\cite{pan2024unifying, luo2023reasoning, hu2023survey, sui2024fidelisfaithfulreasoninglarge}. KGs provide structured, explicit, and often domain-specific factual information, with each fact traceable to its original source~\cite{zheng2023does, agrawal2023hallucinationsurvey}. These properties not only enable verification of the model’s reasoning but also bring transparency to the decision-making process, making KGs a promising avenue for improving the reliability and trustworthiness of LLM outputs. A detailed review of related work is provided in Appendix~\refsec{sec:related_work}.

However, existing benchmarks for evaluating these LLM+KG models predominantly focus on \textbf{closed-ended} tasks~\cite{jin2020medqa,puerto2023metaqa}, where model outputs are restricted to a fixed set of entities, relations~\cite{talmor2019commonsenseqa, 2018openbookqa}, or logical forms~\cite{yih2016webqsp, Talmor2018cwq}. While these benchmarks are useful to measure retrieval accuracy and basic reasoning, such benchmarks fall short in detecting whether a model is \textbf{hallucinating}. In closed-ended settings, errors can stem from incorrect retrieval or fabricated (hallucinated) answers, but conventional metrics like accuracy or precision cannot distinguish between these two failure modes. This limitation becomes problematic for more complex, real-world applications that demand nuanced answers~\cite{kantharaj-etal-2022-opencqa}.

In contrast, our work focuses on open-ended knowledge graph question answering (KGQA), where LLMs are aimed to generate detailed answers that include explicit reasoning paths and supporting facts derived from the KG, as illustrated in Figure~\ref{fig:demo}. This expanded output space offers two key advantages. First, it enables direct measurement of hallucination using metrics such as FActScore~\cite{minetal2023factscore} and SAFE~\cite{wei2024longformfactualitylargelanguage}, which decompose complex responses into atomic statements and verify their factual consistency against external knowledge sources like Wikipedia. Second, longer and more complex answers increase the likelihood of exposing factual errors, aligning with observations from \citet{qiu2024longhalqalongcontexthallucinationevaluation} that hallucinations tend to accumulate in extended responses.
% Thus, while closed-ended settings tend to mask hallucination due to their restrictive nature, our open-ended KGQA more clearly exposes the susceptibility of LLMs to generate hallucinatory content. 
By adopting this open-ended paradigm, we aim to (1) explore whether KGs can enhance the trustworthiness of LLMs in realistic, open-ended scenarios, and (2) provide a comparative analysis to guide the design of methods that leverage KGs to reduce hallucination.

To this end, we introduce \textbf{\underline{O}}pen-ended \textbf{\underline{K}}nowledge-\textbf{\underline{G}}raphs \textbf{\underline{Q}}uestion \textbf{\underline{A}}nswering (\ours{}), a new benchmark tailored to evaluate LLMs augmented with KGs in open-ended QA settings. \ours{} reflects the complexities of practical applications by incorporating diverse question types (see Table~\ref{tab:query_types}), ensuring that queries cannot be answered by simply retrieving isolated KG facts. To simulate real-world conditions where KGs may be \textbf{imperfect} or \textbf{contaminated}—for example, with mislabeled attributes or spurious relations—we propose a benchmark variant, \oursp{} (\refsec{sec:okgqa-p}), which evaluates model robustness when KG semantics and structure are deliberately perturbed. In both settings, we assess hallucination rates alongside overall response quality (details in \refsec{sec:experiment_settings}).

Based on our experiments, we find that (1) integrating KG information generally reduces factual errors, especially for queries requiring deeper reasoning; (2) relying solely on internal LLM reasoning strategies (e.g., Chain-of-Thought~\cite{kim2023cot} and Self-Consistency~\cite{model_self_consistency}) can introduce biases and hallucinations; (3) subgraph-based methods often achieve the best performance for simpler query types; and (4) incorporating KGs effectively reduces hallucinations in LLMs even when the KG is partially contaminated.

\vspace{-1mm}
\section{OKGQA: An Open-ended Knowledge Graph Question-Answering Benchmark}
\label{sec:method}

% \ours{} is a benchmark designed to assess LLMs enhanced with KGs under open-ended, real-world questions answering scenarios. \ours{} is designed to closely reflect the complexities of practical applications using questions of different types and complexities.
% OKGQA evaluates the ability of LLMs+KG methods to utilize and query their integrated knowledge graphs effectively, ensuring a robust test of their semantic understanding and reasoning faculties. This includes managing ambiguous or incomplete information, grasping context, and delivering precise, relevant answers. Overall, OKGQA aims to mirror the unpredictable variety of questions posed in real-world scenarios, pushing the boundaries of what these models can achieve and highlighting areas for future enhancements.
% In this section, we discuss the construction of the dataset, and the construction of \oursp{} which is a variant of \ours{} where the KGs' semantics and structure are deliberately perturbed and contaminated.

\ours{} is a comprehensive benchmark designed to assess how effectively LLMs enhanced with KGs perform in open-ended, real-world-like question answering scenarios. Unlike existing benchmarks that focus primarily on closed-ended tasks, \ours{} presents diverse open-ended question types that mirror the variable nature of practical applications. As illustrated in Figure~\ref{fig:demo}, given a complex query and its corresponding subgraph in a KG, the system must be capable of understanding the relationships within the data and performing human-like reasoning over the KG content to compose a paragraph-long answer. \textbf{In the following section}, we first describe our dataset construction, including query generation via LLM templates and KG subgraph extraction with PPR pruning. We then introduce \oursp{}, a benchmark variant designed to evaluate model robustness under KG perturbations, detailing our perturbation methods and the metrics used to assess semantic and structural deviations. 
Due to page limitations, we provide additional extensions of our benchmark—including a multilingual setup and further analyses—in Appendix~\refsec{sec:extension_of_benchmark}.

\subsection{Dataset Construction}
\label{sec:dataset_construction}

\begin{figure*}[ht]
     \centering
     \begin{subfigure}[b]{0.4\textwidth}
         \centering
         \resizebox{1.1\columnwidth}{!}
         {\begin{tabular}{lcr}
            \toprule
            \textbf{Statistics (on average)} \\
            [2pt]\hdashline\\[-8pt]
            Tokens in query & & 23.97 \\
            Total number of queries && 850 $\rightarrow$ 2,050 \\
            Number of unique DBPedia entities && 816 \\
            \midrule
            \textbf{Before Pruning $\rightarrow$ After PPR Pruning}  && \\
            [2pt]\hdashline\\[-8pt]
            Tokens in subgraph && 348,715 $\rightarrow$ 2,452 \\
            Number of nodes && 7,171 $\rightarrow$ 48\\
            Number of Edges && 8,213 $\rightarrow$ 152 \\
            Avg. Degree && 1.15 $\rightarrow$ 3.17\\
            Clustering Coefficient && 0.00 $\rightarrow$ 0.69\\
            Graph Density && 0.00 $\rightarrow$ 0.07 \\
            \midrule
            \textbf{Query Type} & \textbf{Simple} & \textbf{Complex} \\
                        [2pt]\hdashline\\[-8pt]
            Descriptive &     78 &     11 \\
            Explanatory &     195 &     56 \\
            Predictive &     110  &    55 \\
            Comparative &    72  &      74\\
            Critical &     182 &     17 \\
            Total &     637 &     213 \\
            \bottomrule
        \end{tabular}}
         \caption{\small Dataset statistics and query types}
        \label{tab:dataset_stats_n_ctypes}
     \end{subfigure}
     \hfill
     \hfill
     \hfill
     \hfill 
     \begin{subfigure}[b]{0.55\textwidth}
         \centering
        \includegraphics[width=1\textwidth,trim={0cm 0.2cm 0cm 0cm}]{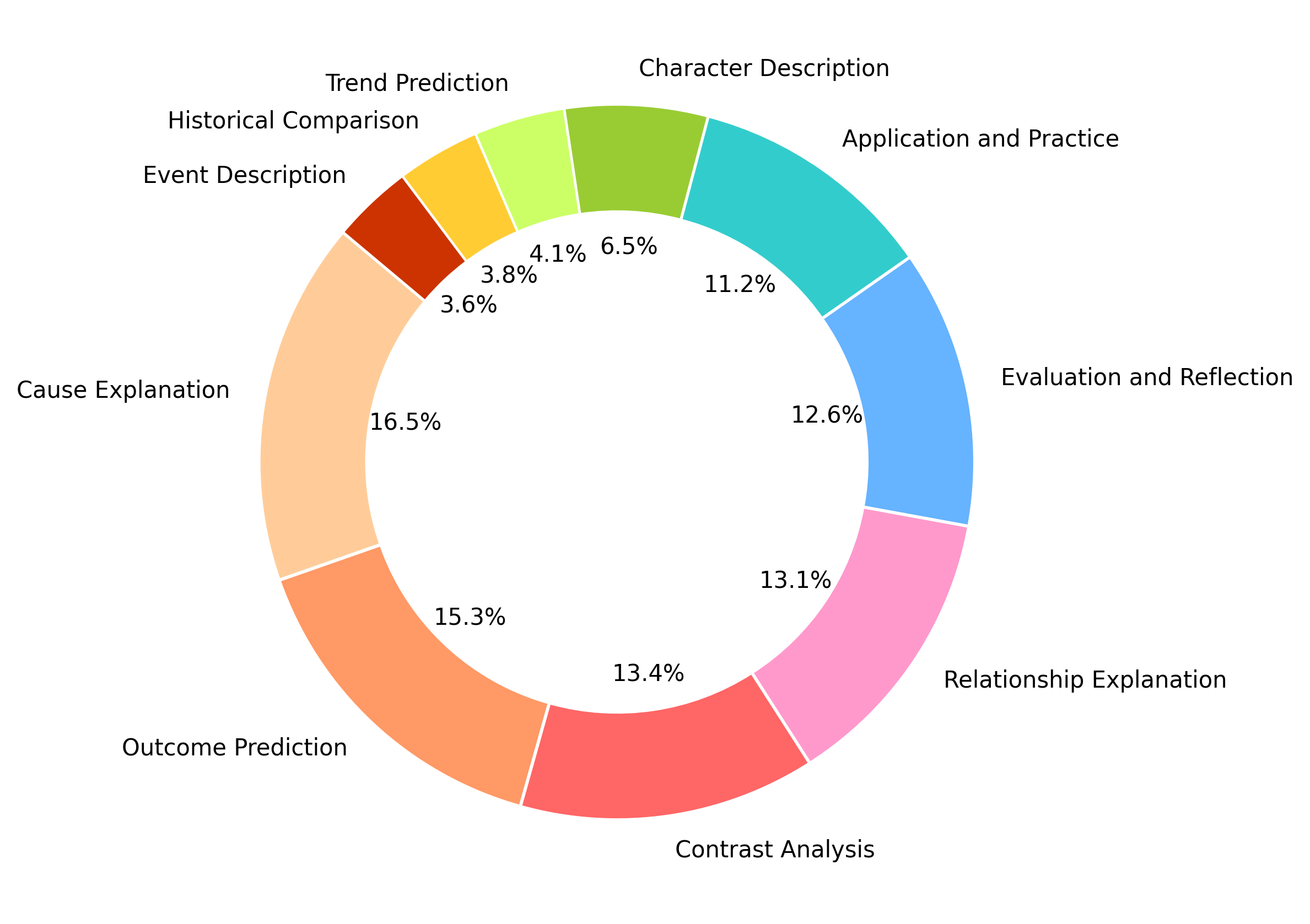}
         \caption{\small Distribution of sub-query types}
        \label{fig:sub-query types}
     \end{subfigure}
     \hfill
     \caption{\small (left) Dataset statistics and query types, (right) Sub-query type distribution}
     \label{fig:dataset_stats_n_ctypes_n_types}
\end{figure*}

\paragraph{Queries.} We utilize a template-based approach to generate a diverse set of queries using LLMs, including categories such as descriptive, explanatory, predictive, comparative, and critical queries. Specific templates and example queries are detailed in Table~\ref{tab:query_types}, with the corresponding prompts provided in Appendix~\ref{sec:prompt_list}.
To ensure that the generated queries reflect real-world complexity and relevance, we adopt an iterative optimization process that combines automated and human evaluations to refine the query generation process (see Appendix~\ref{sec:query_construction} for details). Initially, we generate a diverse set of query candidates from a seed instruction. These candidates are then scored automatically by an LLM-based evaluator, which assigns quality scores $s_\text{auto}$ on a scale of 1-10, where higher scores indicate better performance across multiple criteria. Subsequently, human evaluators assess the same queries, producing normalized scores $s_\text{human}$ within the same range as $s_\text{auto}$. To refine the query generation, we iteratively optimize the input instructions by minimizing the discrepancy between $s_\text{human}$ and $s_\text{auto}$, thereby aligning automated assessments with human judgment. Additionally, the generated queries are categorized by complexity, with detailed statistics shown in Figure~\ref{fig:dataset_stats_n_ctypes_n_types}.

\paragraph{KG Sub-graphs.}
\label{sec:sub_graph_retrieval}
To manage the size of KGs while covering relevant context for the queries, we follow previous work~\cite{yih2016webqsp,Talmor2018cwq} by sampling subgraphs from DBpedia (Noted that all queries in \ours{} can be answered using DBpedia). Specifically, we collect all triples within the $K$-hop neighborhood of entities mentioned in each query. We set $K = 2$ to balance graph coverage and computational feasibility. As increasing beyond $2$-hop subgraphs generally leads to exponential growth in edges and nodes~\cite{jin2020medqa}, which increases excessive noise and complicates graph retrieval\footnote{This choice aligns with common practices in benchmarks such as WebQSP~\cite{yih2016webqsp} and CWQ~\cite{Talmor2018cwq}, where 2-hop subgraphs are widely used for similar KGQA tasks.}. 
To further reduce the size of subgraphs, we leverage Personalized Page-Rank (PPR)~\cite{bahmani2010fastppr} to prune the nodes/edges that are not relevant to the query (the details of the PPR algorithm are discussed in Appendix~\ref{sec:ppr}). We compare the statistics of subgraphs before and after PPR pruning in Figure~\ref{tab:dataset_stats_n_ctypes}.

\subsection{OKGQA-P: Benchmark with Noise \& Perturbations in KGs}
\label{sec:okgqa-p}

KGs are often annotated by humans and can contain errors such as mislabeled attributes or missing relations.
To mimic the real situations where KGs' quality may not be fully reliable, we propose \textbf{\oursp{}} to assess the model performance under deliberately perturbed and contaminated KGs.
We introduce various perturbation scenarios including mislabeled attributes, incorrect relations, and missing connections to test how well models can handle flawed or incomplete KG data. To quantify the degree of perturbation, we measure the semantic and structural similarity between the original and the modified KG as defined below.

% Considering that KGs are typically annotated by humans and are generally accurate and meaningful, we introduce perturbations to edges in the KG to degrade the quality of the KGs, diminishing human comprehensibility. 

% We hypothesis that if original-perturbed KG similarity is low, then a human-like KG-augmented methods should achieve worse performance with the perturbed KG than with the original KG.

\textbf{Notation.} Let $\mathcal{F}_\theta$ be a KG-augmented model, and KG as $\mathcal{G} = (\mathcal{V}, \mathcal{E}, \mathcal{T})$, where $\mathcal{V}$ is the set of entities (nodes), $\mathcal{E}$ is the set of relation types (edges), and $\mathcal{T} = \{(v_1, e, v_2)|v_1, v_2 \in \mathcal{V}, e \in \mathcal{E}\}$ is the set of triplets composed of entities and relations. Let $\mathcal{G}' = (\mathcal{V}, \mathcal{E}', \mathcal{T}')$ be the KG after perturbing $\mathcal{G}$, where $\mathcal{E}' \neq \mathcal{E}$ and $\mathcal{T}' \neq \mathcal{T}$. Let $f(\mathcal{G}, \mathcal{G}')$ be a function that measures the similarity between $\mathcal{G}$ and $\mathcal{G}'$. Let $g(\mathcal{G})$ be the downstream performance when evaluating $\mathcal{F}_\theta$ on data samples $X$ and $\mathcal{G}$.

\begin{table*}[ht]
\centering
\resizebox{0.9\textwidth}{!}{%
\begin{tabular}{@{}ccll@{}}
\toprule
\textbf{Type} & \textbf{Sub-Type} & \multicolumn{1}{c}{\textbf{Description / Template}} & \multicolumn{1}{c}{\textbf{Example}} \\ 
\midrule
\multirow{3}{*}{\begin{tabular}[c]{@{}c@{}}Descriptive\end{tabular}} 
    & Character Description & \begin{tabular}[c]{@{}l@{}}Describe a \highlight{[person]}’s significant contributions \\ during their career.\end{tabular} 
    & \begin{tabular}[c]{@{}l@{}}Please describe \highlightEx{Albert Einstein}’s contributions to the \\ field of \highlightEx{physics}.\end{tabular} \\
\cmidrule(l){2-4} 
    & Event Description & \begin{tabular}[c]{@{}l@{}}Provide a detailed description of the background \\ and course of an \highlight{[event]}.\end{tabular} 
    & \begin{tabular}[c]{@{}l@{}}Please provide a detailed description of the background \\ and course of the \highlightEx{French Revolution}.\end{tabular} \\ 
            [2pt]\hdashline\\[-8pt]
\multirow{3}{*}{\begin{tabular}[c]{@{}c@{}}Explanatory\end{tabular}} 
    & Cause Explanation & \begin{tabular}[c]{@{}l@{}}Why did \highlight{[person]} take \highlight{[action]} at \highlight{[time]}?\end{tabular} 
    & \begin{tabular}[c]{@{}l@{}}Why did \highlightEx{Nixon} choose to \highlightEx{resign} from the presidency \\in \highlightEx{1974}?\end{tabular} \\ 
\cmidrule(l){2-4} 
    & Relationship Explanation & \begin{tabular}[c]{@{}l@{}}Explain the relationship between \highlight{[entity A]} and \\ \highlight{[entity B]} and its significance.\end{tabular} 
    & \begin{tabular}[c]{@{}l@{}}Explain the relationship between \highlightEx{Alexander the Great} \\ and  \highlightEx{Aristotle} and its significance.\end{tabular} \\ 
            [2pt]\hdashline\\[-8pt]
\multirow{3}{*}{\begin{tabular}[c]{@{}c@{}}Predictive\end{tabular}} 
    & Trend Prediction & \begin{tabular}[c]{@{}l@{}}Based on the historical behavior of \highlight{[entity]}, what \\do you think it might do in the future?\end{tabular} 
    & \begin{tabular}[c]{@{}l@{}}Based on \highlightEx{Tesla}’s historical behavior, in which fields do \\ you think it might innovate in the future?\end{tabular} \\ 
\cmidrule(l){2-4} 
    & Outcome Prediction & \begin{tabular}[c]{@{}l@{}}Based on the current situation, how do you predict \\ \highlight{[event]} will develop?\end{tabular} 
    & \begin{tabular}[c]{@{}l@{}}Based on the current international situation, how do you \\ predict \highlightEx{climate change policies} will develop?\end{tabular} \\ 
            [2pt]\hdashline\\[-8pt]
\multirow{3}{*}{\begin{tabular}[c]{@{}c@{}}Comparative\end{tabular}} 
    & Contrast Analysis & \begin{tabular}[c]{@{}l@{}}Compare and contrast the similarities and differences \\between \highlight{[entity A]} and \highlight{[entity B]} in \highlight{[aspect]}.\end{tabular} 
    & \begin{tabular}[c]{@{}l@{}}Compare and contrast the leadership styles of \highlightEx{Steve Jobs} \\ and \highlightEx{Bill Gates}.\end{tabular} \\ 
\cmidrule(l){2-4} 
    & Historical Comparison & \begin{tabular}[c]{@{}l@{}}Compare the impact of \highlight{[historical event A]} and \\\highlight{[historical event B]}.\end{tabular} 
    & \begin{tabular}[c]{@{}l@{}}Compare the impact of \highlightEx{World War I} and \highlightEx{World War II}\\ on the global order.\end{tabular} \\ 
            [2pt]\hdashline\\[-8pt]
\multirow{3}{*}{\begin{tabular}[c]{@{}c@{}}Critical\end{tabular}} 
    & Evaluation and Reflection & \begin{tabular}[c]{@{}l@{}}How do you evaluate the impact of \highlight{[person/event]} \\ on \highlight{[field]}? Please explain your viewpoint.\end{tabular} 
    & \begin{tabular}[c]{@{}l@{}}How do you evaluate \highlightEx{Martin Luther King}’s impact on \\the \highlightEx{civil rights movement}? Please explain your viewpoint.\end{tabular} \\ 
\cmidrule(l){2-4} 
    & Application and Practice & \begin{tabular}[c]{@{}l@{}}How do you think \highlight{[theory/method]} can be applied\\ to \highlight{[practical issue]}? \end{tabular} 
    & \begin{tabular}[c]{@{}l@{}}How do you think \highlightEx{machine learning technology} can be \\applied to \highlightEx{medical diagnostics}?\end{tabular} \\ 
\bottomrule
\end{tabular}}
\caption{\small Query types and examples in OKGQA. \highlight{Brown} is used to highlight the placeholders (\textit{e.g.}, [person], [event]) in description, while \highlightEx{Teal} highlights the specific entities to replace the placeholders. 
% The distribution of the sub-query types can be referred to in Figure~\ref{fig:sub-query types}.
}
\label{tab:query_types}
\end{table*}

\textbf{High-level Procedure.} First, we test $\mathcal{F}_\theta$ on data samples $X$ and $\mathcal{G}$ to get the original performance $g(\mathcal{G})$. Second, we perturb $\mathcal{G}$ to obtain $\mathcal{G}'$. Third, we evaluate $\mathcal{F}_\theta$ on data samples $X$ and $\mathcal{G}'$ to get the perturbed performance $g(\mathcal{G}')$. Finally, we measure $g(\mathcal{G})-g(\mathcal{G}')$ and $f(\mathcal{G}, \mathcal{G}')$ to assess how robust $\mathcal{F}_\theta$ is, \textit{i.e.}, to assess the model performance under conditions where KGs' semantics and structure are deliberately perturbed.

To quantify how much the perturbed KG has deviated from the original KG, \textit{i.e.}, $f(\mathcal{G}, \mathcal{G}')$, we leverage metrics from \cite{raman2020learning} to evaluate semantics (ATS) and structural (SC2D, SD2) similarity between perturbed KG and original KG. Intuitively, ATS leverages a pre-trained LM for link prediction to measure the probability of each edge from $\mathcal{G}'$ existing in $\mathcal{G}$, while SC2D and SD2 measure the structural similarity between two KGs based on local clustering coefficient and degree distribution. For each metric, higher value indicates higher similarity. The detailed description of these metrics can be found in Appendix~\ref{appendix:kg-similarity-metric}, with corresponding results shown in Figure~\ref{fig:ATS_SC2D_SD2}.

For the perturbation methods, we consider four edge-based perturbation heuristics based on \cite{raman2020learning} as follows: 
\begin{itemize}[leftmargin=*]
\setlength\itemsep{0em}
    \item \textbf{Relation Swapping (RS)} randomly chooses two edges from $\mathcal{T}$ and swaps their relations.
    \item \textbf{Relation Replacement (RR)} randomly chooses an edge $(v_1, e, v_2)\in\mathcal{T}$, and replaces the $e_1$ with another relation $e_2 = \mathrm{argmin}_{e\in\mathcal{E}}S_\mathcal{G}(v_1, e, v_2)$, where $S_\mathcal{G}(\cdot)$ is a KG score function adapted from ATS. This yield \enquote{harder negatives} - triplets that are semantically similar but incorrect.
    % The replacement is performed such that the resulting $(v_1, e_2, v_2)$ has the closest score to the original triplet $(v_1, e_1, v_2)$ to generate \enquote{harder negatives} - triplets that are semantically similar but incorrect, thereby challenging the model to different subtle distinctions. Noted that $S_{G}$ differs from the original ATS, as it operates on individual triplets rather than comparing two entire KGs.
    \item \textbf{Edge Rewiring (ER)} randomly chooses an edge $(v_1, e, v_2)\in\mathcal{T}$, then replaces $v_2$ with another entity $v_3 \in\mathcal{E} \backslash \mathcal{N}_1(v_1)$, where $\mathcal{N}_1(v_1)$ represents the 1-hop neighborhood of $v_1$.
    \item \textbf{Edge Deletion (ED)} randomly chooses an edge $(v_1, e, v_2) \in\mathcal{T}$ and deletes it.
\end{itemize}

We control perturbation level by adjusting the percentage of edges in $\mathcal{G}$ that are perturbed. Refer to Figures~\ref{fig:ATS_SC2D_SD2} and \ref{fig:performance_okgqap} for empirical results.

\section{Exploring KG-augmented framework for Reducing Hallucination}
\label{sec:framework}
\begin{figure}[t]
    \centering
    \includegraphics[width=\linewidth]{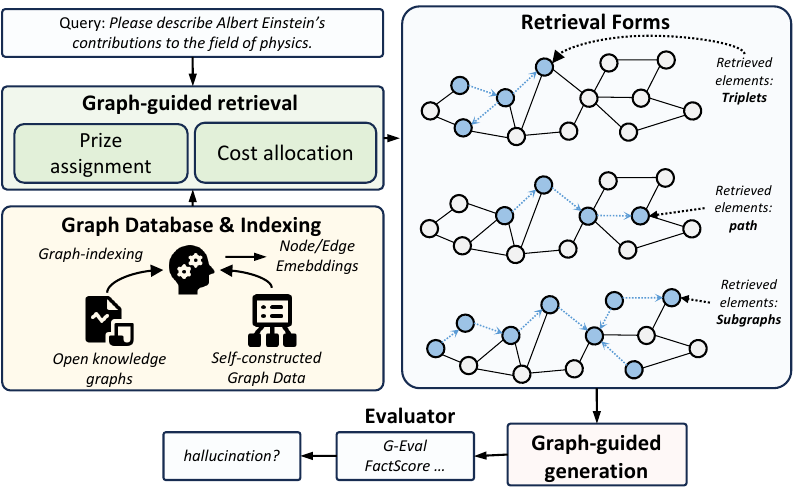}
    \caption{\small Overview of KG-augmented framework.}
    \vspace{-4mm}
    \label{fig:framework}
\end{figure}

To explore whether KG-augmented approaches can mitigate LLMs' hallucination, we propose a unified framework as shown in Figure~\ref{fig:framework}. Our framework follows the paradigm of retrieval augmented generation (RAG)~\cite{edge2024graphrag,baeketal2023kaping}, which retrieves essential information from the KGs, and then uses the retrieved knowledge to enhance the LLM's generation (\refsec{sec:formalization}). It consists of two components, \textit{i.e.}, \emph{Graph-guided retrieval} (\refsec{sec:g-retrieval}) and \emph{Graph-guided generator} (\refsec{sec:g-generator}), with a variety of algorithmic design choices. We analyze the strategies within each component in \refsec{sec:experiments}, aiming to shed light on the best practices for leveraging KGs for reducing hallucinations in LLMs.

\subsection{Formalization}
\label{sec:formalization}

We formalize the KG-augmented framework as follows. Given a user query $q$, a pretrained language model generates a paragraph-like answer $a$ by modeling the conditional probability $p(a \vert q)$. To explore whether KGs help reduce hallucinations of LLMs, we introduce the retrieved knowledge $\mathcal{Z}$ from the KG and define:
\begin{equation}
\label{Eq:frame}
\setlength{\abovedisplayskip}{4pt} % Space above equations
\setlength{\belowdisplayskip}{2pt} % Space below equations
p(a \vert q) = \sum_{\mathcal{Z} \subseteq \mathcal{G}} p_{\phi}(a \vert q, \mathcal{Z}) p_{\theta}(\mathcal{Z} \vert q, \mathcal{G}),
\end{equation}
where $p_{\phi}(a \vert q, \mathcal{Z})$ is the likelihood of generating the paragraph-like answer $a$ conditioned on 
$q$ and $\mathcal{Z}$ (parameterized by $\phi$), and $p_{\theta}(\mathcal{Z} \vert q, \mathcal{G})$ models the retrieval of knowledge subsets (parameterized by $\theta$).
Because the number of possible subsets 
$\mathcal{Z}$ can be exponentially large relative to the size of $\mathcal{G}$, we approximate the sum by selecting the most probable knowledge subset: $\mathcal{Z}^*= \text{argmax}_{\mathcal{Z}\in\mathcal{G}}p_{\theta}(\mathcal{Z}|q, \mathcal{G})$, yielding:
\begin{equation}
\setlength{\abovedisplayskip}{4pt} % Space above equations
\setlength{\belowdisplayskip}{2pt} % Space below equations
    p(a \vert q) \approx p_{\phi}(a \vert q, \mathcal{Z}^*) p_{\theta}(\mathcal{Z}^* \vert q, \mathcal{G})
\end{equation}
% This approach identifies the most relevant subset $\mathcal{Z}^*$ for query $q$ and then conditions the generation of $a$ on both $q$ and $\mathcal{Z}^*$.
% In this framework, the retrieval module identifies the optimal subset $\mathcal{Z}^*$ from $\mathcal{G}$ that is most relevant to the query 
% $q$, and the generation module then produces a comprehensive, paragraph-like response conditioned on both $q$ and $\mathcal{Z}^*$.

\subsection{Graph-guided retrieval (G-retrieval)}
\label{sec:g-retrieval}

Our goal in G-retrieval is to extract a compact yet informative subset $\mathcal{Z}^*$ from the KG that best supports answering the user query $q$. We first encode the query and all KG elements (nodes/edges) into a unified embedding space using a language model. We then measure the relevance of each element to $q$ (e.g., via cosine similarity) and identify a set of top-$k$ nodes and edges for the query. 

To balance retrieving as many relevant nodes and edges as possible while keeping the $\mathcal{Z}^*$ size manageable, we adopt a \textbf{prize-cost trade-off strategy}~\cite{balasprize} to guide the retrieval process:
(1) \textit{Prize assignment}: based on the computed similarity scores, we assign prizes to nodes and edges to quantify their relevance to the query. Specifically, we assign the top-$k$ nodes/edges with descending prize values from $k$ to 1, while nodes and edges outside the top-$k$ receive a prize of 0. Formally:  \( p_v = \max(0, k - \text{rank}(v) + 1) \) and \( p_e = \max(0, k - \text{rank}(e) + 1) \). (2) \textit{Cost allocation}: to manage the retrieved knowledge size, we assign penalties as cost $C_e$ during the expansion of the retrieved paths or subgraphs. The final retrieval process aims to maximize the total prize (i.e., relevance) while minimizing associated costs.

We explore three retrieval variants for G-retrieval design (e.g., triplets, paths and subgraphs) as demonstrated in Figure~\ref{fig:framework}. 
\begin{itemize}[leftmargin=*]
\setlength\itemsep{0em}
    \item \textbf{Triplet-retrieval}: retrieves a fixed number of triplets with the highest total prize assigned to their respective triplets.
    \item \textbf{Path-retrieval}: starting from a fixed number of $k$ of high-prize nodes, we greedily expand paths $\mathcal{P} = \{v_1, e_1, v_2, \dots, e_{n-1}, v_n\}$ to maximize score: $S(\mathcal{P}) = \sum_{i=1}^{n} p_{v_i} + \sum_{i=1}^{n-1} p_{e_i} - \sum_{i=1}^{n-1} c_e$. We use a priority queue to iteratively return paths with top-scores and subject to maximum lengths and cycles. The details of path-retrieval can be found in Appendix~\ref{sec:path-retrieval}.
    \item \textbf{Sub-graph retrieval}: building on \citet{he2024g-retriever}, we use the Prize-Collecting Steiner Tree (PCST) algorithm to find a connected subgraph $\mathcal{S}$ that maximizes $S(\mathcal{S}) = \sum_{n \in V_{\mathcal{S}}} p_{v_i} + \sum_{e \in E_{\mathcal{S}}} p_{e_i} - \sum_{e in E_{\mathcal{S}}} c_e$. Unlike in path-retrieval, we only yield one subgraph that maximizes the total score.
\end{itemize}

\subsection{Graph-guided Generation (G-Generator)}
\label{sec:g-generator}
After retrieving $\mathcal{Z}^*$, the G-Generator use this knowledge to generate the paragraph-like response the user query. The generation is modeled as a sequential decision-making process: at each time step $t$, token $a_t$ is generated conditioned on $q$, $\mathcal{Z}^*$, and the previously generated tokens $a_{0:t-1}$:
\begin{equation}
\setlength{\abovedisplayskip}{0pt} % Space above equations
\setlength{\belowdisplayskip}{0pt} % Space below equations
p(a|q,\mathcal{Z}^*) = \prod_{t=1}^T p_\theta(a_t \vert q, \mathcal{Z}^*, a_{0: t-1}),
\end{equation}
where $\theta$ denotes the parameters of a neural text generation model. The generation stops when an end-of-sequence token is produced or when the maximum sequence length $T$ is reached.

\section{Experiments}
\label{sec:experiments}

In this section, we first introduce the evaluation metrics and experiment setup, and then focus on two main research questions and provide relevant analysis: RQ1: Can KGs reduce hallucination in LLMs? and RQ2: How are KG-Aware methods affected by noise/perturbations in KGs?

\subsection{Evaluation Metrics \& Setup}
\label{sec:experiment_settings}

We quantify LLM hallucinations using two public metrics: \textbf{FActScore}~\citep{minetal2023factscore} and \textbf{SAFE}~\citep{wei2024longformfactualitylargelanguage}. 
\textbf{FActScore} measures factual precision by decomposing a long-form text into atomic facts and validating each against a reliable knowledge base like Wikipedia. In contrast, \textbf{SAFE} employs a language model as an investigative agent that iteratively employs Google Search queries to assess whether search results support the fact. For both metrics, we report the proportion of supported atomic facts out of the total atomic facts extracted from LLM responses.

In addition to the hallucination metrics, we propose four metrics using LLM-as-evaluator~\citep{li2024llmsasjudges} to quantify the quality of generated responses from LLM~\citep{edge2024graphrag, wang2023chatgpt}. In specific, we use G-Eval~\citep{liu2023gval} framework for the evaluation and provide relevant Wikipedia pages of each query as context to enhance G-Eval's robustness and stability. The four metrics are defined as follows: 
(1) \textbf{Context Relevance}: measures how well the generated response aligns with the provided context.
(2) \textbf{Comprehensiveness}: assesses how thoroughly the answer addresses all aspects and details of the question.
(3) \textbf{Correctness}: measures the clarity and specificity of the generated answer to the question.
(4) \textbf{Empowerment}: evaluates how well the generated answer helps the reader understand the topic and make informed decisions.
The corresponding prompts are detailed in Appendix~\ref{sec:prompt_list}.

We use gpt-4o-mini (from November 2024 to January 2025) as LLM backbone for all the evaluation metrics. As using LLM-as-evaluator frameworks may raise concern regarding \textbf{potential self-enhancement} or bias from the selection of the backbone models~\cite{gu2024survey,li2024llmsasjudges}, we conduct additional analysis in Appendix~\ref{sec:llm_evaluation_clarity} (including human evaluation alignment and cross-validation across different LLM backbones), and find that the choice of LLMs in the LLM-as-evaluator framework has little impact on the overall evaluation and the results demonstrate high correlation with the human evaluation, supporting the reliability of our testing.

For our experiments, we consider a range of widely used LLMs of different scales for testing, including GPT-4o, GPT-4o-mini (from November 2024 to January 2025), Llama-3.1-8B-instruct~\citep{dubey2024llama3_1}, Mistral-7B-instruct-v0.3~\citep{jiang2023mistral}, and Gemma-2-9B-it~\citep{gemmateam2024gemma2improvingopen}. Considering the trade-off between cost and performance, we use text-embedding-3-small model from OpenAI (from November 2024 to January 2025) as embedding model for G-retrieval process. To ensure the reproducibility of the experiments, we set $\mathrm{temperature}=0.7$ and $\mathrm{top\_p}=1.0$ for all models. We use the API service from OpenAI\footnote{\url{https://openai.com/}} and OpenRouter\footnote{\url{https://openrouter.ai/}} for our experiments which host detailed snapshots of the utilized model versions.

\begin{table*}[t]
\centering
\resizebox{0.97\textwidth}{!}{
\begin{tabular}{@{}lcccccc@{}}
\toprule
\multicolumn{1}{c}{\multirow{3}{*}{Models}} & \multicolumn{4}{c}{\textbf{G-Eval}} & \multicolumn{2}{c}{\textbf{Hallucination}} \\ 
\cmidrule(l){2-5} \cmidrule(l){6-7} 
\multicolumn{1}{c}{}  & Context Relevance  & Comprehensiveness & Correctness  & Empowerment    & SAFE   & FActScore    \\ \midrule
\multicolumn{7}{c}{\textbf{Baseline: Without External Knowledge (Zero-shot prompting)}}  \\
\rowcolor{gray!20}
GPT-4o  & $68.12\%\pm0.88\%$  & $65.41\%\pm0.79\%$  & $60.41\%\pm0.38\%$    & $62.41\%\pm0.84\%$    & $82.47\%\pm0.62\%$   & $55.34\%\pm0.93\%$  \\
GPT-4o-mini  & $63.21\%\pm0.49\%$  & $60.11\%\pm0.47\%$  & $55.43\%\pm0.63\%$    & $58.72\%\pm0.62\%$  & $80.14\%\pm0.89\%$  & $50.23\%\pm1.01\%$  \\
llama-3.1-8b-instruct  & $57.12\%\pm0.91\%$   & $54.74\%\pm1.20\%$    & $49.01\%\pm0.61\%$   & $52.21\%\pm0.71\%$   & $79.33\%\pm0.91\%$  & $45.14\%\pm0.32\%$\\ 
mistral-7B-Instruct-v0.3  & $55.71\%\pm1.21\%$  & $52.00\%\pm1.31\%$  & $47.03\%\pm0.94\%$   & $50.13\%\pm1.04\%$   & $78.27\%\pm0.83\%$  & $44.37\%\pm1.23\%$\\ 
gemma-2-9b-it & $53.63\%\pm1.33\%$   & $50.00\%\pm1.33\%$    & $45.72\%\pm0.71\%$   & $48.15\%\pm0.93\%$   & $77.11\%\pm0.78\%$  & $40.94\%\pm0.83\%$\\ \midrule

\multicolumn{7}{c}{\textbf{Baseline: Without External Knowledge (4-shot prompting)}}  \\
\rowcolor{gray!20}
GPT-4o  & $70.61\%\pm0.62\%$  & $67.43\%\pm0.81\%$  & $62.33\%\pm0.37\%$    & $64.51\%\pm0.12\%$    & $83.39\%\pm0.53\%$   & $57.45\%\pm0.78\%$  \\
GPT-4o-mini  & $65.53\%\pm0.94\%$  & $62.33\%\pm1.03\%$  & $57.23\%\pm0.68\%$    & $60.47\%\pm0.83\%$  & $81.62\%\pm0.69\%$  & $52.34\%\pm0.76\%$  \\
llama-3.1-8b-instruct  & $59.43\%\pm0.32\%$   & $56.31\%\pm0.78\%$    & $51.27\%\pm0.32\%$   & $54.33\%\pm0.41\%$   & $80.27\%\pm0.78\%$  & $47.24\%\pm1.03\%$\\
mistral-7B-Instruct-v0.3  & $57.34\%\pm1.04\%$  & $54.13\%\pm1.31\%$  & $49.27\%\pm0.84\%$   & $52.46\%\pm0.94\%$   & $79.12\%\pm0.87\%$  & $45.13\%\pm1.42\%$\\
gemma-2-9b-it & $55.24\%\pm1.49\%$   & $52.27\%\pm1.21\%$    & $47.14\%\pm0.36\%$   & $50.36\%\pm0.51\%$   & $78.00\%\pm0.77\%$  & $44.32\%\pm1.58\%$\\ \midrule

\multicolumn{7}{c}{\textbf{Baseline: With Wikipedia documents}}  \\
\rowcolor{gray!20}
GPT-4o - IRCoT  & $73.12\%\pm0.32\%$  & $69.23\%\pm0.42\%$  & $66.33\%\pm0.34\%$    & $65.51\%\pm0.11\%$    & $87.39\%\pm0.68\%$   & $69.45\%\pm0.34\%$  \\
GPT-4o-mini - IRCoT & $70.31\%\pm0.32\%$  & $64.42\%\pm1.31\%$  & $61.37\%\pm0.48\%$    & $63.89\%\pm0.72\%$  & $84.72\%\pm0.48\%$  & $65.72\%\pm1.03\%$  \\ \midrule

\multicolumn{7}{c}{\textbf{Var-1: With CoT Prompting}}  \\
\rowcolor{gray!20}
GPT-4o - CoT  & $72.76\%\pm0.92\%$  & $69.56\%\pm0.74\%$  & $64.48\%\pm0.63\%$    & $66.69\%\pm0.69\%$    & $80.07\%\pm0.83\%$   & $54.30\%\pm0.87\%$  \\
GPT-4o - CoT+SC  & $75.81\%\pm0.65\%$  & $71.62\%\pm0.74\%$  & $66.55\%\pm0.75\%$    & $68.74\%\pm0.15\%$    & $79.03\%\pm0.48\%$   & $53.23\%\pm0.78\%$  \\
% llama-3.1-8b-instruct - CoT  & $61.54\%\pm0.95\%$   & $58.35\%\pm1.05\%$    & $53.31\%\pm0.71\%$   & $56.42\%\pm0.83\%$   & $77.07\%\pm0.85\%$  & $46.15\%\pm0.54\%$\\
llama-3.1-8b-instruct - CoT+SC & $63.69\%\pm0.32\%$   & $60.44\%\pm0.59\%$    & $55.46\%\pm0.52\%$   & $58.53\%\pm1.11\%$   & $76.00\%\pm0.63\%$  & $45.05\%\pm0.97\%$\\
% mistral-7B-Instruct-v0.3 - CoT  & $59.58\%\pm0.43\%$   & $56.23\%\pm2.31\%$    & $51.28\%\pm1.31\%$   & $54.33\%\pm0.72\%$   & $75.04\%\pm0.95\%$  & $43.03\%\pm1.03\%$\\
mistral-7B-Instruct-v0.3 - CoT+SC & $61.35\%\pm0.93\%$   & $58.33\%\pm1.02\%$    & $53.42\%\pm0.79\%$   & $56.47\%\pm0.85\%$   & $74.30\%\pm0.21\%$  & $42.00\%\pm0.29\%$\\
% gemma-2-9b-it - CoT  & $57.34\%\pm1.05\%$  & $54.12\%\pm0.32\%$  & $49.27\%\pm0.85\%$   & $52.12\%\pm0.95\%$   & $72.07\%\pm1.05\%$  & $40.13\%\pm0.49\%$\\
gemma-2-9b-it - CoT+SC & $59.42\%\pm0.27\%$  & $56.27\%\pm0.84\%$  & $51.34\%\pm1.42\%$   & $54.34\%\pm1.31\%$   & $71.09\%\pm0.43\%$  & $39.85\%\pm1.03\%$\\ \midrule

\multicolumn{7}{c}{\textbf{Var-2: With Triplets Extracted from KGs Provided}}  \\
\rowcolor{gray!20}
GPT-4o  & 
$74.62\%\pm0.65\%$  & $70.44\%\pm0.79\%$  & $65.37\%\pm0.72\%$    & $67.12\%\pm0.71\%$    & $89.20\%\pm1.42\%$   & $72.53\%\pm0.83\%$  \\
GPT-4o-mini  & $69.50\%\pm0.81\%$  & $65.03\%\pm0.92\%$  & $60.21\%\pm0.65\%$    & $63.43\%\pm1.01\%$  & $87.52\%\pm0.34\%$  & $67.73\%\pm0.95\%$  \\
llama-3.1-8b-instruct  & $63.45\%\pm1.13\%$   & $59.33\%\pm1.05\%$    & $54.23\%\pm0.75\%$   & $57.33\%\pm0.12\%$   & $85.37\%\pm0.72\%$  & $62.37\%\pm0.82\%$\\
mistral-7B-Instruct-v0.3  & $61.34\%\pm0.31\%$  & $57.21\%\pm0.89\%$  & $52.29\%\pm0.32\%$   & $55.12\%\pm0.43\%$   & $84.21\%\pm0.84\%$  & $60.28\%\pm1.05\%$\\
gemma-2-9b-it & $59.25\%\pm1.06\%$  & $55.29\%\pm0.44\%$  & $50.15\%\pm0.85\%$   & $53.73\%\pm0.95\%$   & $83.18\%\pm0.43\%$  & $58.13\%\pm0.91\%$\\ 
GPT-4o - CoT+SC  & $76.71\%\pm0.53\%$  & $72.34\%\pm0.21\%$  & $67.33\%\pm1.31\%$    & $69.64\%\pm0.33\%$    & $88.11\%\pm0.57\%$   & $71.45\%\pm0.53\%$  \\ \midrule

\multicolumn{7}{c}{\textbf{Var-3: With Paths Extracted from KGs Provided}}  \\
\rowcolor{gray!20}
GPT-4o  & 
$78.71\%\pm0.53\%$  & $74.53\%\pm0.31\%$  & $69.42\%\pm0.23\%$    & $71.63\%\pm0.61\%$    & $90.20\%\pm0.59\%$   & $\textbf{75.61\%}\pm\textbf{0.51\%}$  \\
GPT-4o-mini  & $73.64\%\pm0.93\%$  & $69.41\%\pm0.22\%$  & $64.35\%\pm0.72\%$    & $67.52\%\pm0.82\%$  & $88.22\%\pm0.34\%$  & $70.53\%\pm0.24\%$  \\
llama-3.1-8b-instruct  & $67.51\%\pm0.46\%$   & $63.62\%\pm1.39\%$    & $58.41\%\pm0.93\%$   & $61.57\%\pm0.94\%$   & $86.33\%\pm0.94\%$  & $65.42\%\pm0.95\%$ \\
mistral-7B-Instruct-v0.3  & $65.48\%\pm0.94\%$   & $61.37\%\pm1.01\%$    & $56.34\%\pm0.23\%$   & $59.45\%\pm0.43\%$   & $85.26\%\pm0.85\%$  & $63.31\%\pm1.33\%$\\
gemma-2-9b-it & $63.35\%\pm1.37\%$  & $59.23\%\pm0.91\%$  & $54.31\%\pm0.91\%$   & $57.41\%\pm0.27\%$   & $84.13\%\pm0.21\%$  & $61.23\%\pm1.04\%$\\ 
GPT-4o - CoT+SC  & $80.87\%\pm0.42\%$  & $76.60\%\pm0.65\%$  & $71.54\%\pm0.53\%$    & $73.79\%\pm1.21\%$    & $89.11\%\pm0.63\%$   & $74.53\%\pm0.24\%$  \\ \midrule

\multicolumn{7}{c}{\textbf{Var-4: With Subgraphs Extracted from KGs Provided}}  \\
\rowcolor{gray!20}
GPT-4o  & $80.81\%\pm0.43\%$  & $76.63\%\pm0.65\%$  & $71.57\%\pm0.51\%$    & $73.70\%\pm0.62\%$    & $\textbf{90.83\%}\pm\textbf{0.63\%}$   & $75.33\%\pm0.29\%$  \\
GPT-4o-mini  & $75.70\%\pm0.44\%$  & $71.51\%\pm0.83\%$  & $66.43\%\pm0.76\%$    & $69.60\%\pm0.65\%$  & $88.71\%\pm0.72\%$  & $70.12\%\pm0.87\%$  \\
llama-3.1-8b-instruct  & $69.61\%\pm0.84\%$   & $65.45\%\pm0.93\%$    & $60.41\%\pm0.65\%$   & $63.42\%\pm0.45\%$   & $86.12\%\pm0.35\%$  & $65.44\%\pm0.87\%$\\ 
mistral-7B-Instruct-v0.3  & $67.55\%\pm0.87\%$   & $63.35\%\pm0.43\%$    & $58.37\%\pm0.71\%$   & $61.45\%\pm0.32\%$   & $85.21\%\pm0.81\%$  & $63.12\%\pm0.94\%$\\ 
gemma-2-9b-it & $65.45\%\pm0.95\%$  & $61.23\%\pm1.0\%$  & $56.31\%\pm0.35\%$   & $59.40\%\pm0.85\%$   & $84.51\%\pm0.99\%$  & $63.74\%\pm0.49\%$\\ 
GPT-4o - CoT+SC  & $\textbf{82.90\%}\pm\textbf{0.57\%}$  & $\textbf{78.72\%}\pm\textbf{0.61\%}$  & $\textbf{73.64\%}\pm\textbf{0.43\%}$    & $\textbf{75.80\%}\pm\textbf{0.75\%}$    & $89.12\%\pm0.94\%$   & $75.42\%\pm1.31\%$  \\ \bottomrule
\end{tabular}}
\caption{Comparison results of various forms of information extracted from the KGs.}
\label{tab:comparaison}
\end{table*}

\subsection{RQ1: Main Results - Can KGs Reduce Hallucination in LLMs?}

To explore whether KGs can help reduce hallucination in LLMs, we benchmark the LLMs in different settings. We use zero-shot and few-shot prompting as baselines without injecting external knowledge. In addition, we consider leveraging LLMs' internal knowledge to do Chain-of-thought~\citep{kim2023cot}, or self-consistency~\citep{model_self_consistency}, and more general RAG systems like IRCoT~\citep{trivedi2022interleaving} which retrieves paragraphs from Wikipedia to augment CoT generation. For LLMs augmented with KGs, we consider three KG retrieval variants: triplets, paths, and subgraphs to study the impact of G-retrieval for reducing LLMs' hallucinations. The results are shown in Table~\ref{tab:comparaison} and Figure~\ref{fig:comparison}. We obtain some intriguing findings:

\begin{figure}[ht]
\centering
\includegraphics[width=\linewidth]{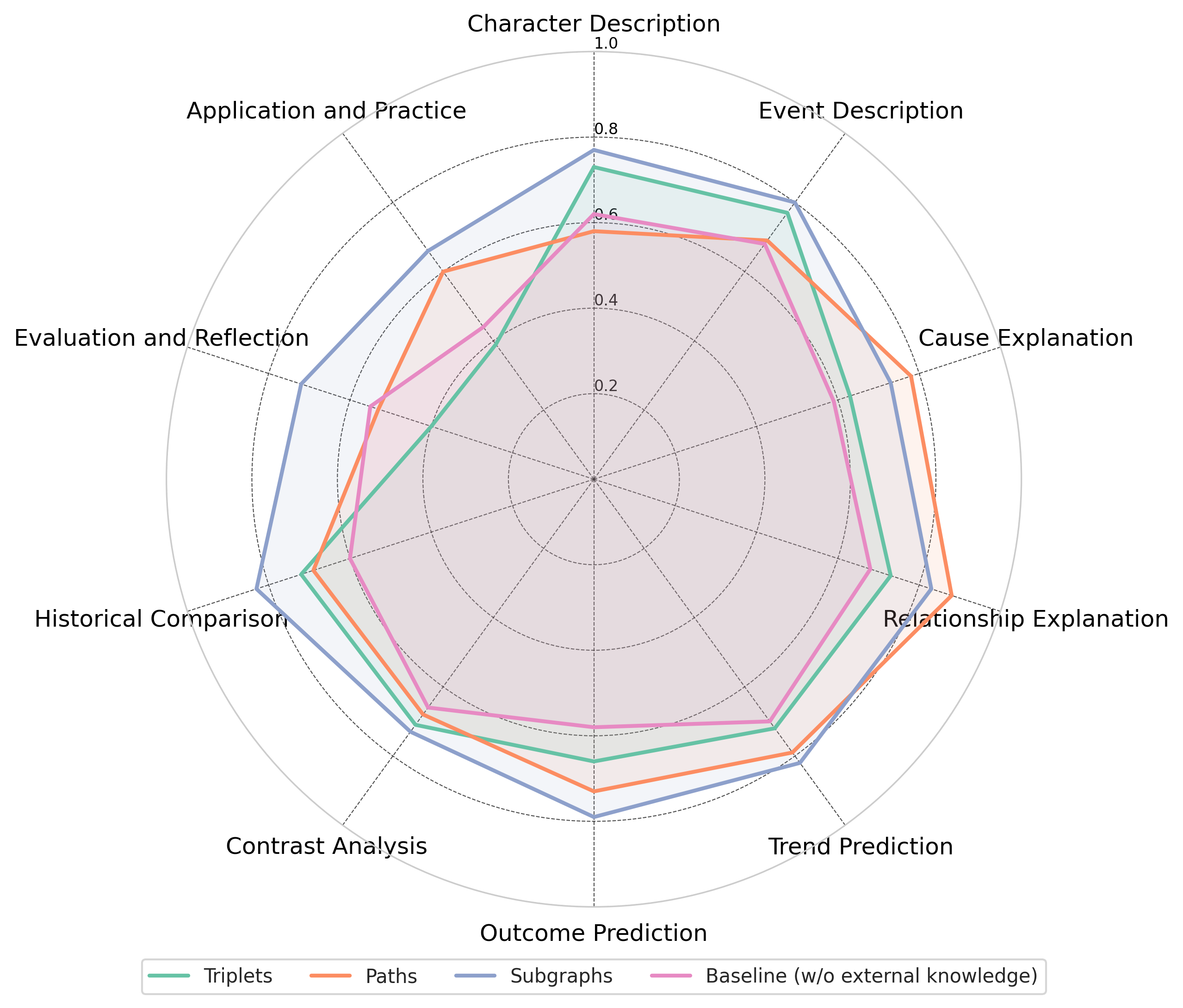}
\caption{\small Comparison results of different forms of information over different queries.}
% \vspace{-4mm}
\label{fig:comparison}
\end{figure}

\textbf{Retrieving KG information can indeed mitigate factual errors in the responses.}
Methods integrating knowledge extracted from KGs show clear improvements in factual accuracy and comprehension scores compared to the baselines. For example, under Var-2 (triplet retrieval), GPT-4o achieves a FActScore of 72.55\% $\pm$ 0.85\%, which is a significant increase over the baseline score of 55.35\% $\pm$0.95\%. Moreover, these methods can be combined with strategies like CoT+SC, enhancing response quality with minimal increase in hallucination ratio. The radar chart in Figure~\ref{fig:comparison} further emphasizes that in most query types, integrating knowledge retrieved from KGs mitigates the hallucination issue compared to baselines, particularly in query types such as \enquote{Evaluation and Reflection,} \enquote{Outcome Prediction,} and \enquote{Cause Explanation,} which require more reasoning and analysis rather than merely listing information. The findings also apply to open-source models like mistral-7B-Instruct-v0.3 and Llama-3.1-8B-instruct, illustrating the consistency of the finding. In addition, compared with RAG method IRCoT~\cite{trivedi2022ircot}, leveraging Wikipedia documents, improves performance over zero-shot and 4-shot prompting by providing broad contextual support, it struggles with correctness and hallucination control due to the potential introduction of irrelevant or conflicting information. Our KG-based methods consistently outperform IRCoT, particularly in correctness, SAFE, and FActScore.

\textbf{Directly performing reasoning in the LLM itself does not mitigate hallucinations.}
We benchmark the hallucination ratio of LLMs using internal reasoning strategies like CoT and Self-consistency. As shown in Var-1 in Table~\ref{tab:comparaison}, these methods can improve response quality (i.e., G-Eval) compared to baselines, but do not consistently improve factuality, and sometimes even diminish. This shows that relying solely on internal reasoning is inadequate for mitigating hallucinations, highlighting the necessity for external knowledge to address this issue effectively.

\begin{figure*}[t]
% \vspace{-4mm}
    \centering
    % Subfigure 1: ED
    \begin{subfigure}[b]{0.24\textwidth}
        \centering
        \includegraphics[width=\textwidth]{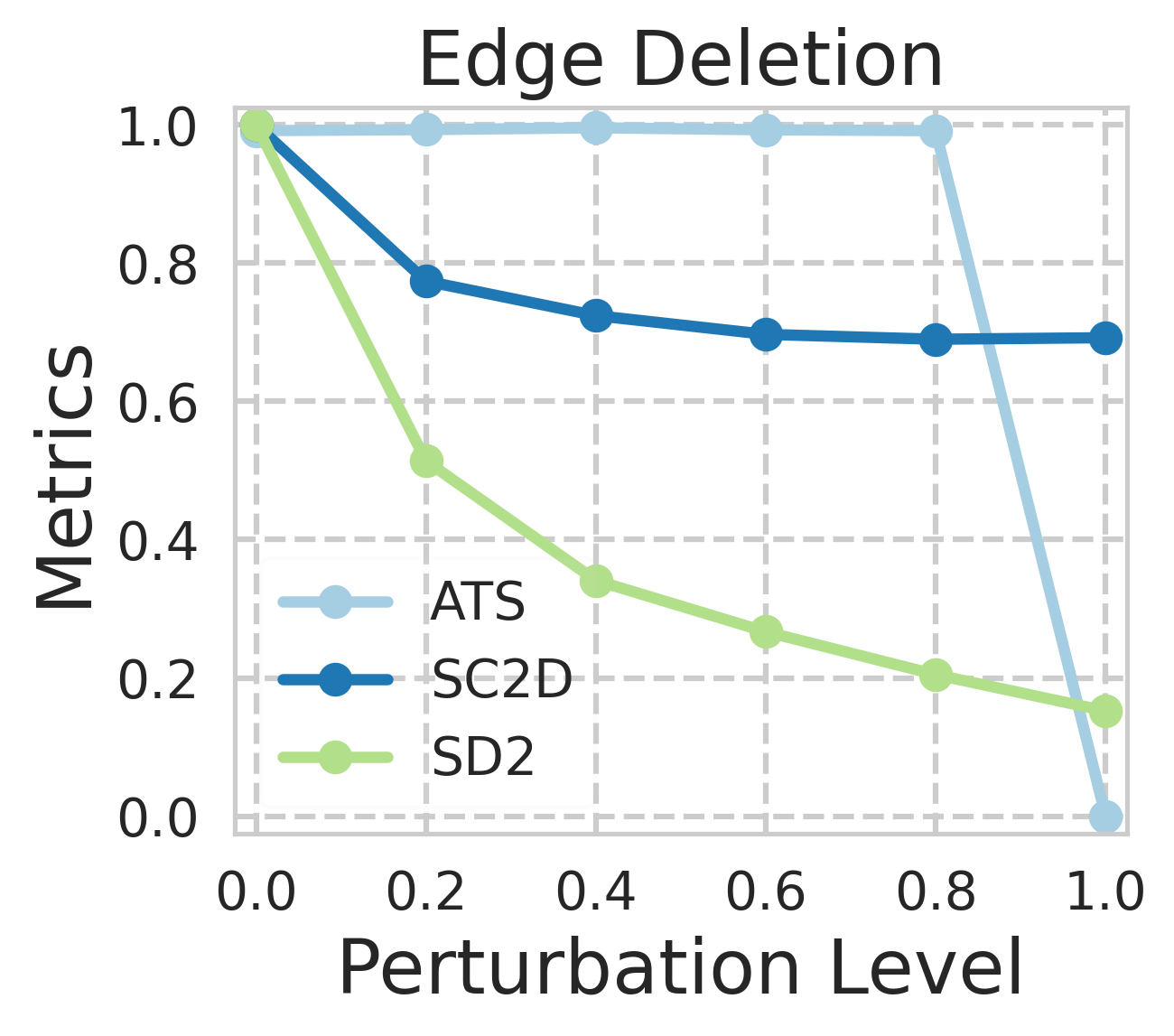}
        \caption{Edge Deletion}
        \label{fig:ED}
    \end{subfigure}
    \hfill
    % Subfigure 2: ER
    \begin{subfigure}[b]{0.24\textwidth}
        \centering
        \includegraphics[width=\textwidth]{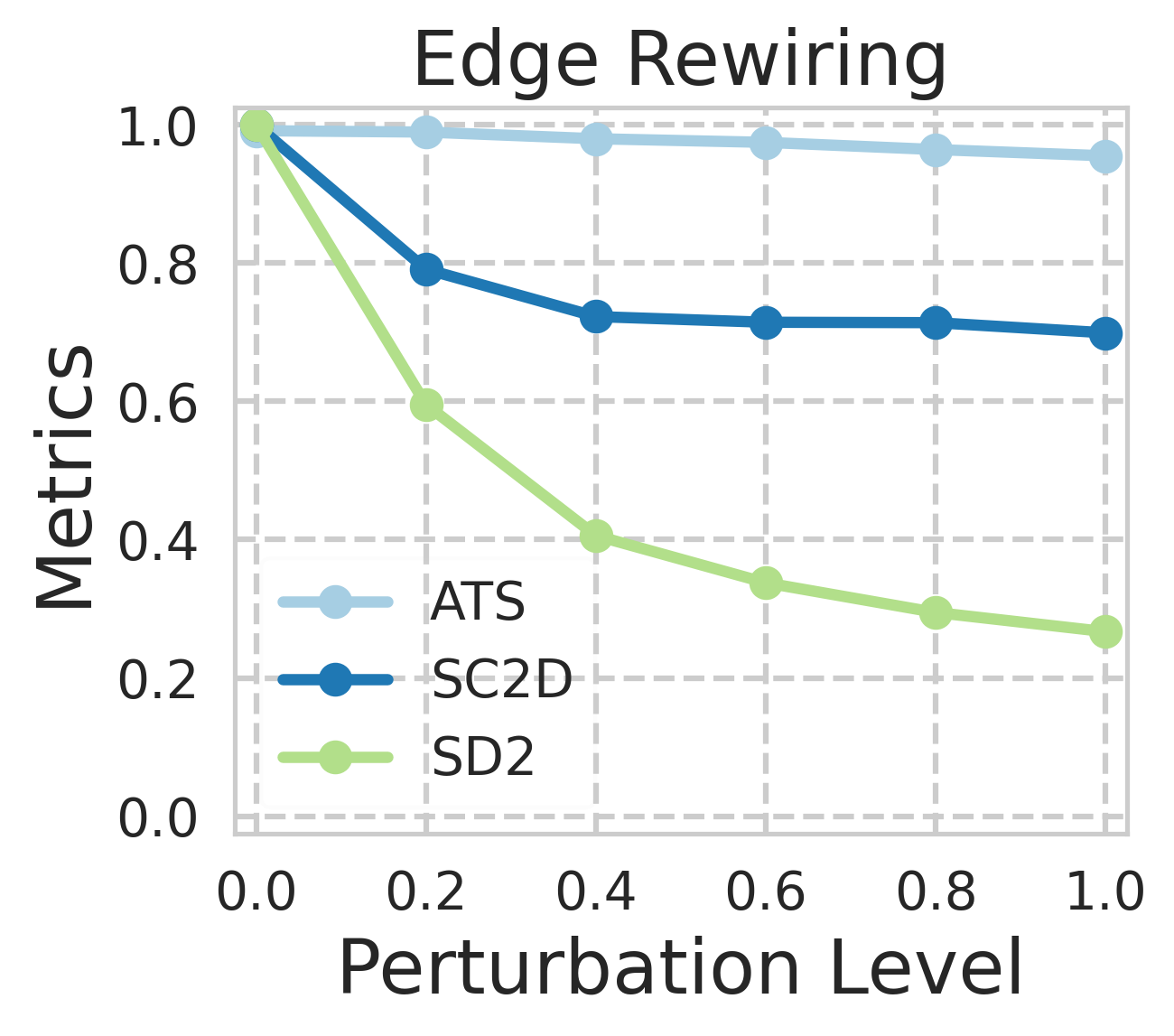}
        \caption{Edge Rewiring}
        \label{fig:ER}
    \end{subfigure}
    \hfill
    % Subfigure 3: RR
    \begin{subfigure}[b]{0.24\textwidth}
        \centering
        \includegraphics[width=\textwidth]{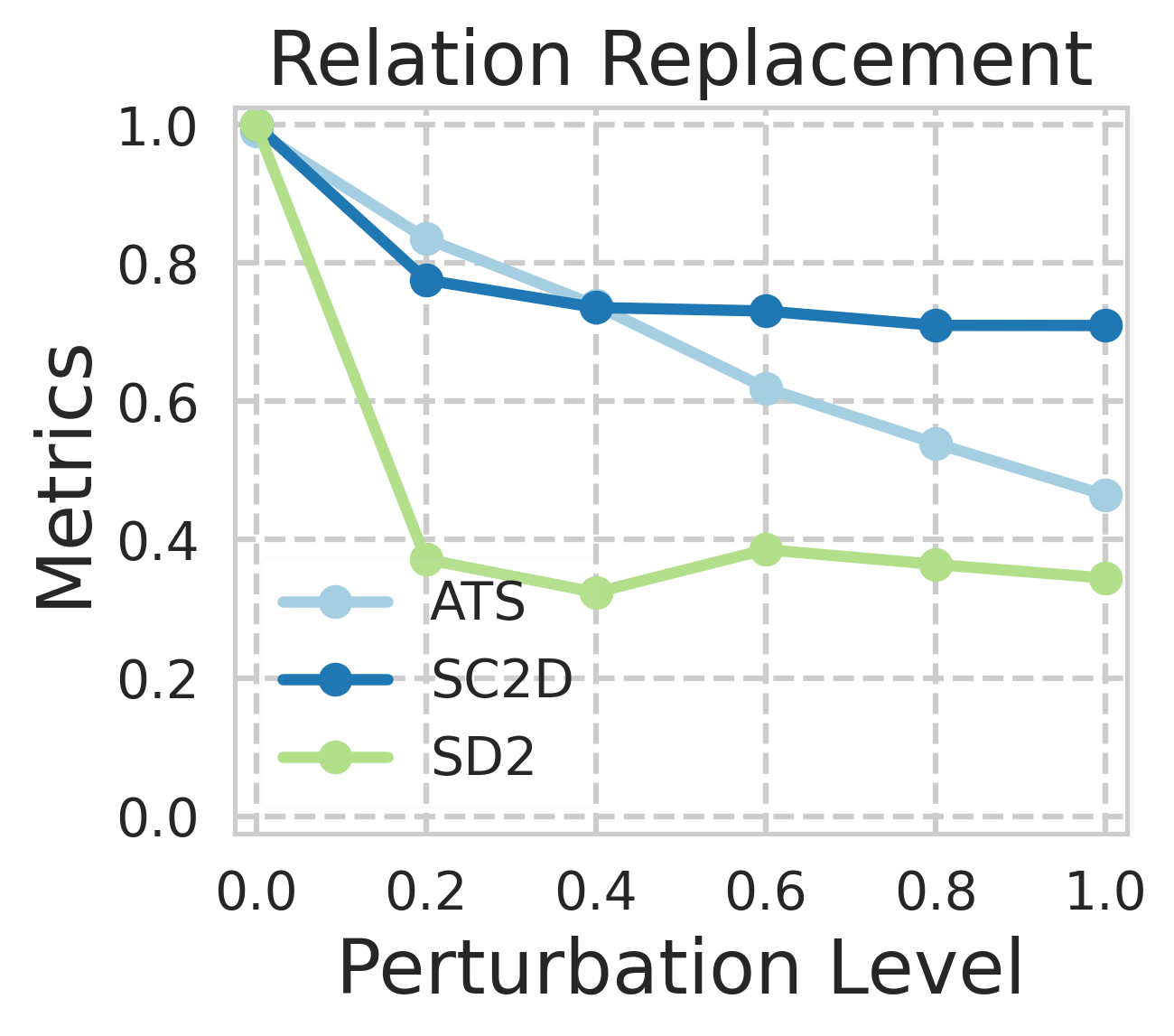}
        \caption{Relation Replacement}
        \label{fig:RR}
    \end{subfigure}
    \hfill
    % Subfigure 4: RS
    \begin{subfigure}[b]{0.24\textwidth}
        \centering
        \includegraphics[width=\textwidth]{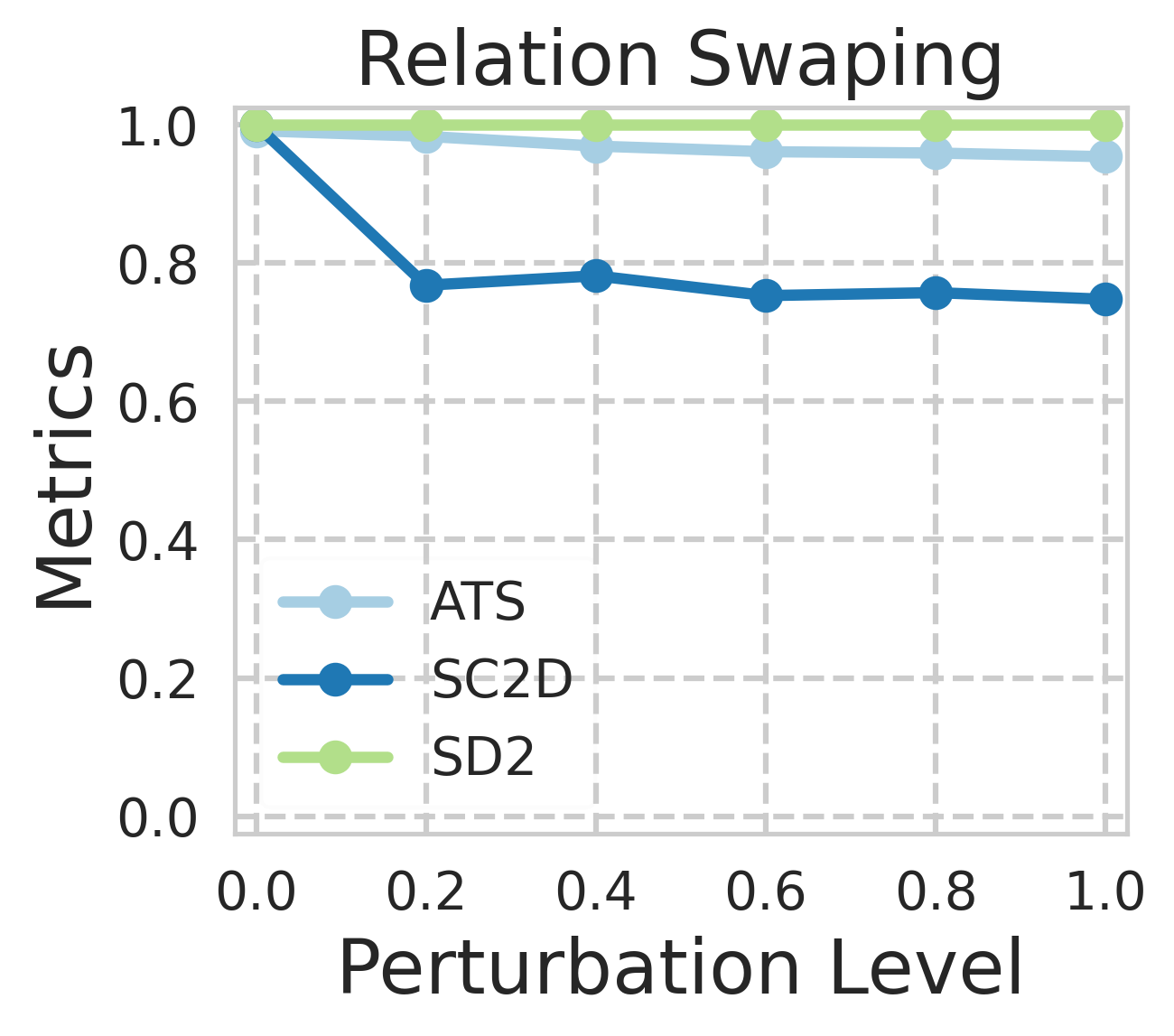}
        \caption{Relation Swapping}
        \label{fig:RS}
    \end{subfigure}
    % Main caption for all subfigures
    \caption{\small Performance Metrics (ATS, SC2D, SD2) vs. Perturbation Level for Different Perturbation Methods.}
    \label{fig:ATS_SC2D_SD2}
\end{figure*}

\begin{figure*}[t]
% \vspace{-4mm}
    \centering
    % Subfigure 1: ED
    \begin{subfigure}[b]{0.24\textwidth}
        \centering
        \includegraphics[width=\textwidth]{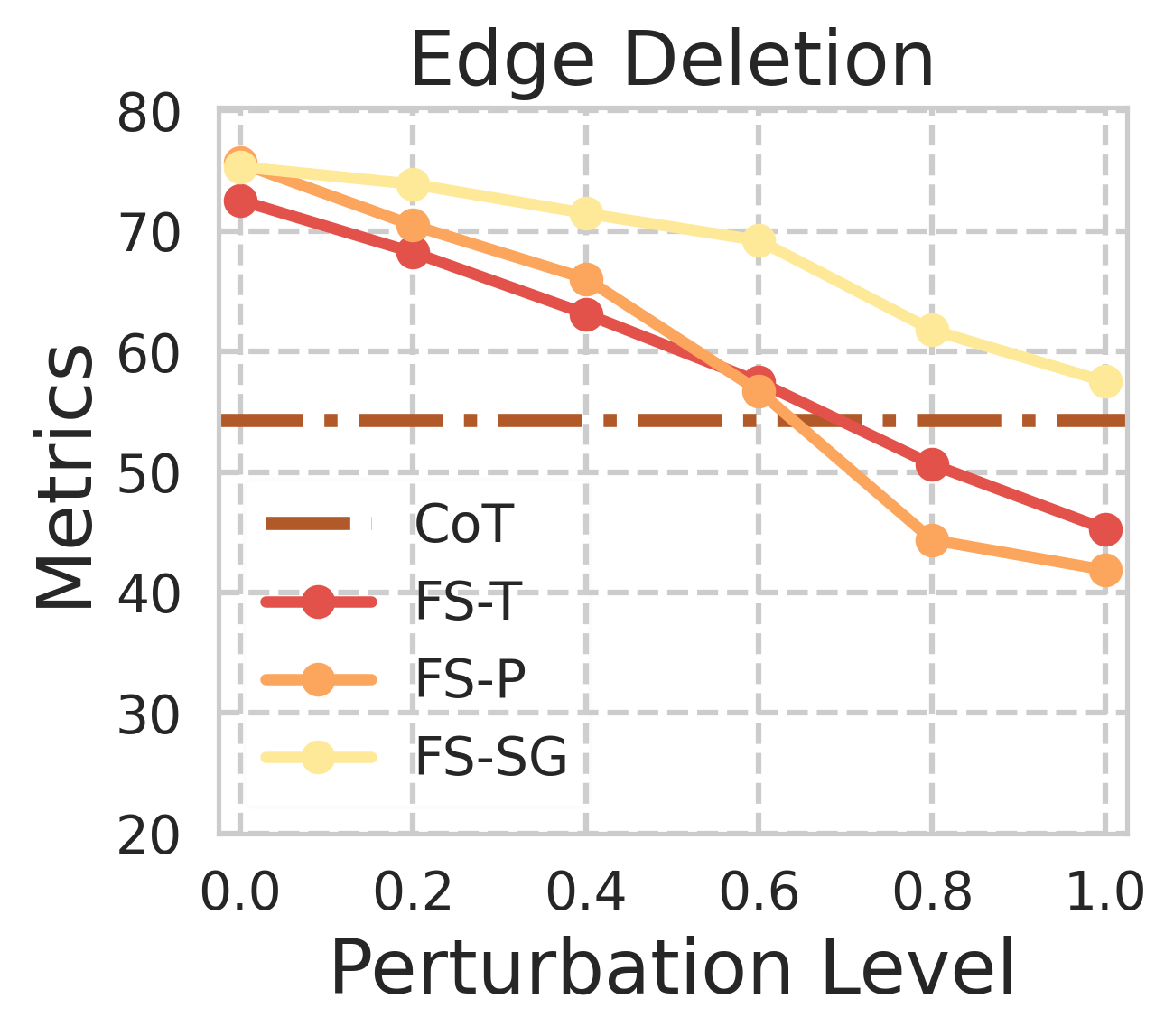}
        \caption{Edge Deletion}
        \label{fig:ED_x}
    \end{subfigure}
    \hfill
    % Subfigure 2: ER
    \begin{subfigure}[b]{0.24\textwidth}
        \centering
        \includegraphics[width=\textwidth]{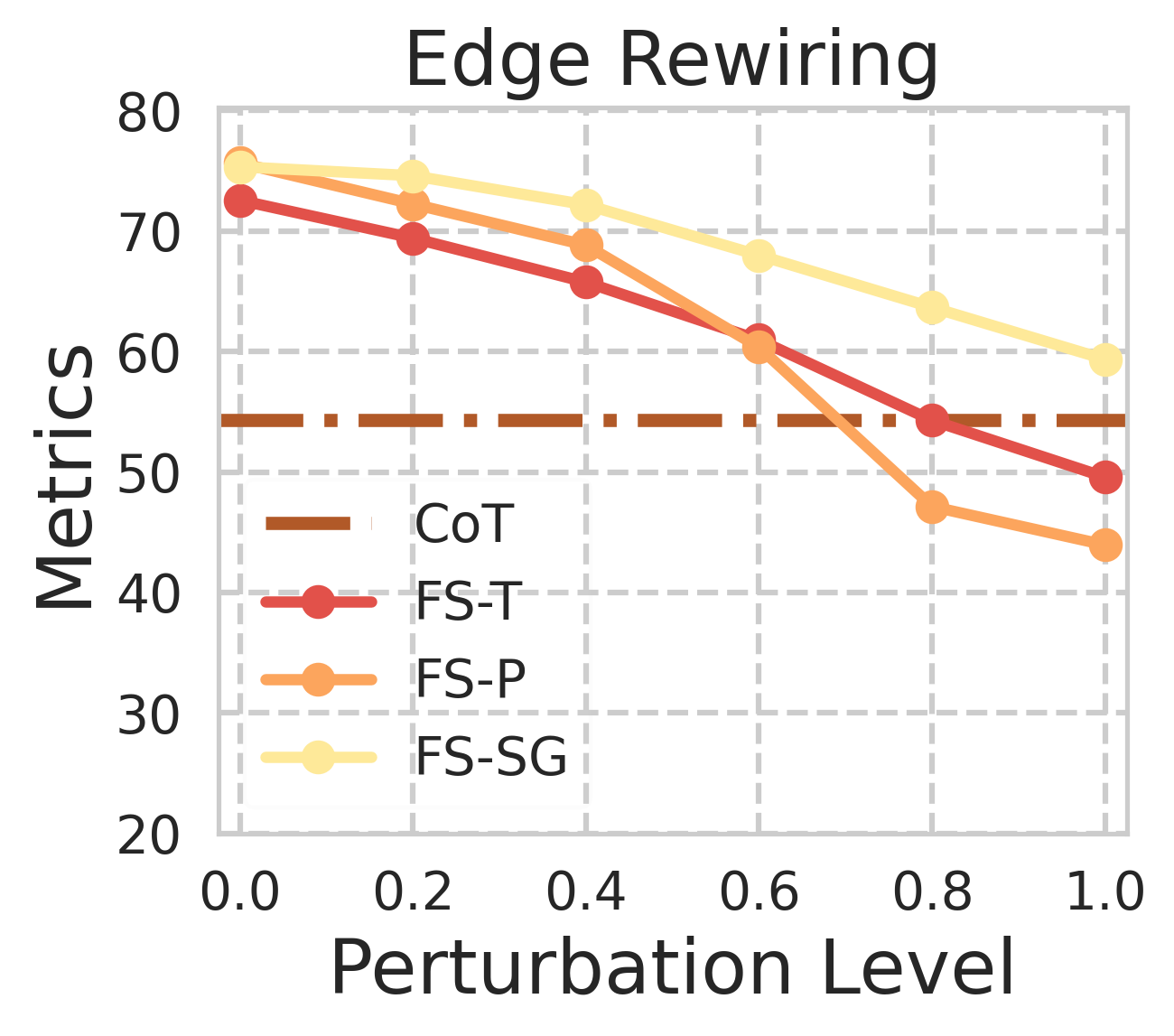}
        \caption{Edge Rewiring}
        \label{fig:ER_x}
    \end{subfigure}
    \hfill
    % Subfigure 3: RR
    \begin{subfigure}[b]{0.24\textwidth}
        \centering
        \includegraphics[width=\textwidth]{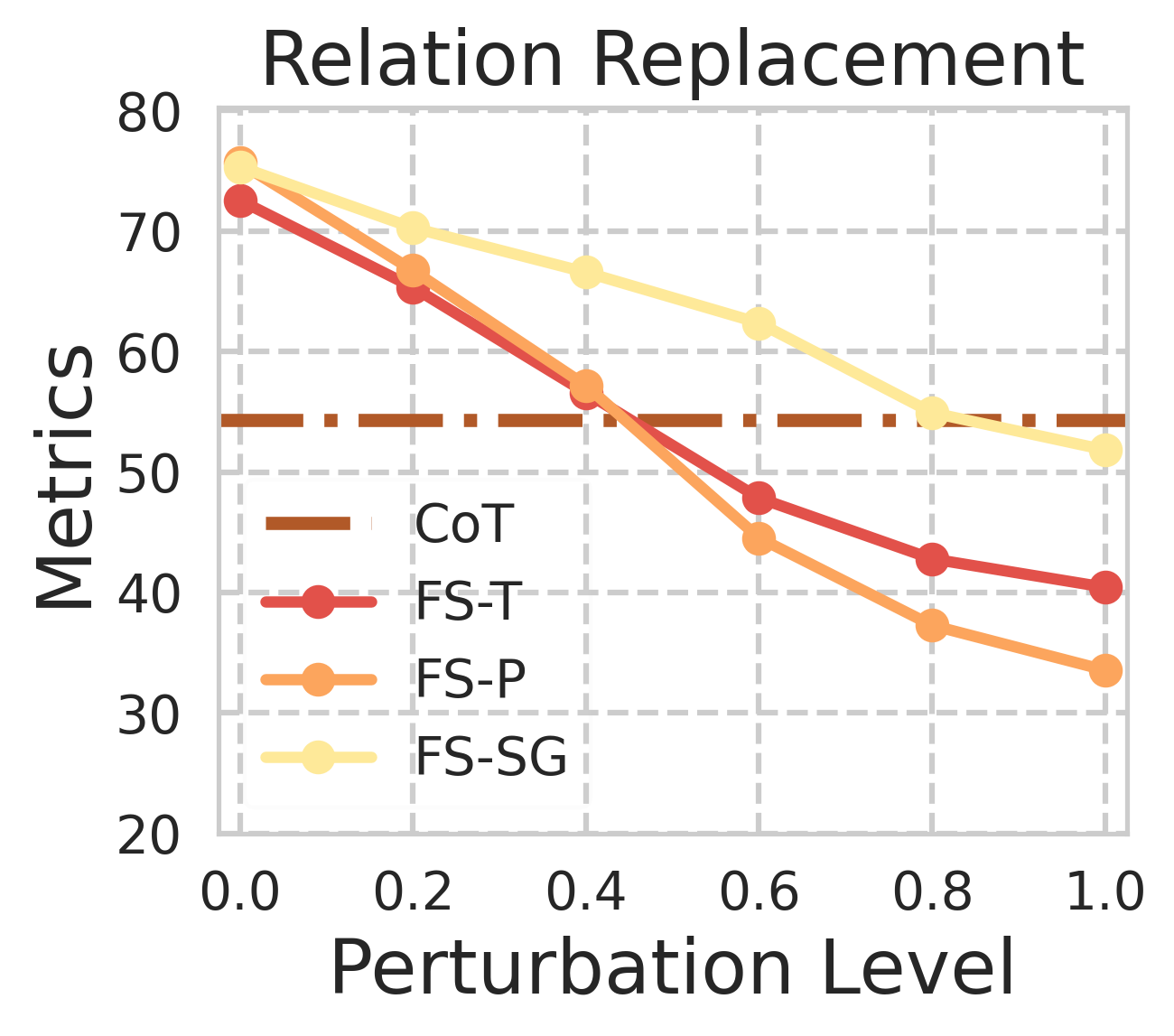}
        \caption{Relation Replacement}
        \label{fig:RR_x}
    \end{subfigure}
    \hfill
    % Subfigure 4: RS
    \begin{subfigure}[b]{0.24\textwidth}
        \centering
        \includegraphics[width=\textwidth]{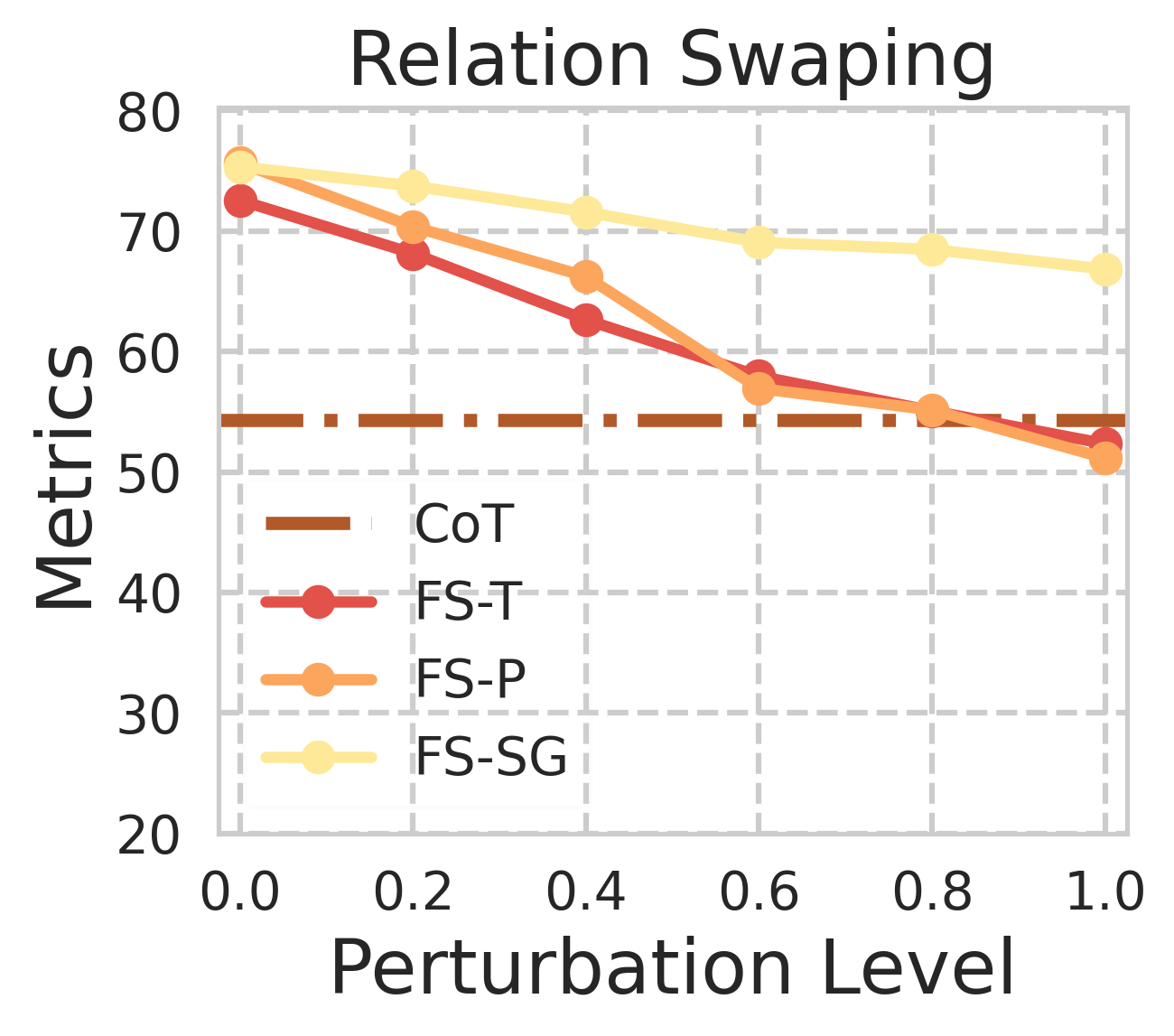}
        \caption{Relation Swapping}
        \label{fig:RS_x}
    \end{subfigure}
    % Main caption for all subfigures
    \caption{\small Performance Metric (FActScore) vs. Perturbation Level for Different Perturbation Methods and Different Retrieval Methods. \textbf{FS-T} refers to FActScore metric using triplets, \textbf{FS-P} refers to using paths, and \textbf{FS-SG} refers to using sub-graphs.}
    \label{fig:performance_okgqap}
\end{figure*}

\textbf{Subgraph retrieval generally achieves best performance across different query types, especially for simpler queries.}
We demonstrate the performance of different retrieval methods across different query types in Figure~\ref{fig:comparison}, showing that subgraphs achieve the best performance. Especially for simpler queries (\enquote{Character Description} and \enquote{Event Description} which do not require intensive reasoning). Even for queries like \enquote{Relationship Explanation} and \enquote{Cause Explanation} which require stepwise reasoning, subgraph methods still demonstrate promising performance.
This suggests that while different forms of retrieved knowledge offer unique benefits for specific types of queries, subgraphs provide consistently good performance.

\subsection{RQ2: How Are KG-Aware Methods Affected by Noise / Perturbations in KGs?}

We test different KG-augmented LLMs on our \oursp{} setting, where we deliberately perturb and contaminate the semantics and structure of KGs to simulate the real-world situation where KGs may not have high quality. Specifically, we consider different perturbation methods discussed in \refsec{sec:okgqa-p} and control the perturbation level based on the percentage of KG edges being perturbed. We first illustrate how much the perturbed KG has been deviated from the original KG with the increase of perturbation level, shown in Figure~\ref{fig:ATS_SC2D_SD2}. It shows that the perturbation methods like edge deletion, rewiring and swapping have relatively weak influence on ATS (which intuitively measures semantic similarity), even as the perturbation level increases. For the edge deletion methods, only if the perturbation level reaches 1.0, the ATS goes to 0, otherwise, the ATS remains higher compared to other settings. 

Figure~\ref{fig:performance_okgqap} illustrates the hallucination ratio using different methods on the perturbed KGs. We observe that (1) FS-SG consistently outperforms FS-T and FS-P even at higher perturbation levels, demonstrating its robustness by maintaining higher scores as perturbations increase. (2) FS-T and FS-P exhibit similar trends, each showing a significant performance drop as perturbation levels increase. Particularly, performance of FS-T and FS-P deteriorate when the perturbation level reaches 50\%, \textit{i.e.}, becoming worse than the baseline using CoT. (3) On the setting using Relation Replacement which severely harms the semantics of the KGs, FS-T and FS-P decline more sharply than FS-SG. However, it still outperforms the baseline when the perturbation level is smaller than 40\%. 

\textbf{In summary}, we find that the effectiveness of KG-derived information diminishes with a perturbation level at 50\%, surpassing this level leads to a further decrease in performance. We think that before this perturbation level at 50\%, incorporating external knowledge from KGs can mitigate hallucinations in LLMs compared to baseline using CoT. Considering practical scenarios, platforms like Wikidata are less likely to have perturbations as severe as 50\% due to their ongoing updates and community-based quality control. This ensures the relevance and applicability of our findings in real-world settings.

\section{Conclusion}
\label{sec:conclusion}

In this paper, we propose \ours{} and variant \oursp{}, to assess LLMs enhanced with KGs under open-ended, real-world question answering scenarios. Unlike existing benchmarks that focus primarily on closed ended tasks, OKGQA presents diverse open-ended question types that mirror the unpredictable nature of practical applications. We conduct a series of experiments and analyze the effectiveness of various retrieval methods and LLMs of different magnitudes, providing insights for further research. Our results underscore the significance of integrating KGs with LLMs to help reduce hallucination of LLMs, even in circumstances where the KGs are  contaminated.

% This benchmark aims to advance the field by providing a standardized way to evaluate LLM+KG systems' trustworthiness and reasoning capabilities, ultimately helping identify promising directions for improving these hybrid systems in real-world applications. 

% \section*{Acknowledgments}

\section{Limitations}

Our proposed benchmark primarily use DBpedia as the knowledge source, which may not generalize well to testing scenarios requiring highly specialized or domain-specific knowledge. Testing domain-specific open-ended QA may require constructing sub-graphs from domain-specific KGs. In addition, the study assumes a static KG for reasoning and retrieval. In dynamic environments where knowledge is continuously updated, maintaining and integrating real-time changes remains a challenge and may requires further design.

\bibliography{ref}

\clearpage
\appendix

\section{Implementation Details}
\subsection{Query Construction}
\label{sec:query_construction}

In this section, we discuss the details of the query construction of \ours{}. We first introduce the human-in-the-loop process to optimize the seed instruction for generating the queries, as shown in \refsec{sec:human_in_the_loop}. We then present the metrics to quantify the generated queries in \refsec{sec:metrics_for_generated_queries}. Subsequently, we provide experiments results of human-in-the-loop process and demonstrate the Pearson correlation coefficients between human and LLM scores across different rounds of optimization, and verify the inter-rater reliability across different LLM evaluators in \refsec{sec:verifying_human_in_the_loop}.

\subsubsection{Human-in-the-loop for instruction optimization}
\label{sec:human_in_the_loop}

\begin{figure}[ht]
    \centering
    \includegraphics[width=1\linewidth]{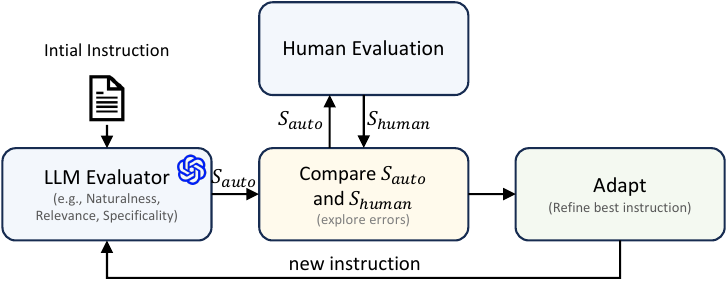}
    \caption{\small Human-in-the-loop of query construction.}
    \label{fig:human_in_loop}
\end{figure}

To ensure that the generated queries represent real-world scenarios and complexities, we propose a human-in-the-loop process to optimize the seed instruction used for generation, as shown in Figure~\ref{fig:human_in_loop}. 
To ensure clarity, we summarize this optimization process here: 
\begin{itemize}
\setlength\itemsep{0em}
\item Step 1: Generate a set of queries from an initial instruction.
\item Step 2: Collect automatic evaluation scores $s_\mathbf{auto}$ by LLMs and human-label scores $s_\mathbf{human}$ by human annotators for these queries (normalized to the same range).
\item Step 3: Identify patterns of discrepancies between these scores.
\item Step 4: Let the LLM analysis the identified patterns to generate new instructions,
\end{itemize}

Steps 3 and 4 are performed by prompting the LLM with the instruction specified in \refsec{prompt:instruction_tuner}, and the entire process from steps 1 to 4 is iterated to minimize the discrepancy between $s_\mathbf{auto}$ and $s_\mathbf{human}$. This procedure closely resembles the way of reinforcement learning with human feedback (RLHF)~\citep{ouyang2022rlhf} and inherits the benefit that labeling the reward of the LLMs' output is much easier than directly labeling the outputs.

\subsubsection{Metrics for generated queries}
\label{sec:metrics_for_generated_queries}

We consider five metrics to measure the quality of the generated queries: (1) \textbf{Naturalness}: assessing how fluid and human-like the query sounds; (2) \textbf{Relevance}: measuring whether the query pertains directly to the entity and the context provided; (3) \textbf{Specificity}: determining the level of detail and granularity included in the query, ensuring it is not too broad or vague; (4) \textbf{Novelty}: evaluating the uniqueness of the query, ensuring it is not just a repetitive or common question; (5) \textbf{Actionability}: gauging whether the query prompts clear, definite answers or actions that are feasible within the given context. Each of these angles contributes to a holistic evaluation of the query's effectiveness and relevance in real-world applications.

\subsubsection{Verifying human-in-the-loop}
\label{sec:verifying_human_in_the_loop}

For the human-label scores $s_\mathbf{human}$ collection, 
we have three evaluators participating in the manual assessment of query quality. All of the evaluators are computer science majors with fluent English skills. As the evaluation centers on various linguistic metrics such as naturalness, relevance, specificity, novelty, and actionability, we only require the evaluators to possess a fundamental understanding of English without restricting their majors. We calculate the Pearson correlation coefficients between human and LLM scores as shown in Table~\ref{tab:metrics_rounds}. It shows that as the rounds progress, agreement between humans and LLMs increases, suggesting that iterative feedback improves alignment between human annotation and LLM responses.

\begin{table}[ht]
\centering
\resizebox{\linewidth}{!}{
\begin{tabular}{lcccc}
\toprule
\textbf{Metric}      & \textbf{Round 1} & \textbf{Round 2} & \textbf{Round 3} & \textbf{Round 4} \\ 
\midrule
Naturalness          & 0.60             & 0.65             & 0.69             & 0.74             \\
Relevance            & 0.55             & 0.59             & 0.64             & 0.70             \\
Specificity          & 0.46             & 0.54             & 0.60             & 0.65             \\
Novelty              & 0.49             & 0.57             & 0.63             & 0.67             \\
Actionability        & 0.33             & 0.41             & 0.48             & 0.53             \\ 
\bottomrule
\end{tabular}}
\caption{Pearson correlation coefficients between human and LLM scores across rounds.}
\label{tab:metrics_rounds}
\end{table}

In addition, we also consider verifying the inter-rater reliability across three evaluators as shown in Table~\ref{tab:evaluator_agreement}. We report the Cohen's Kappa coefficient for each pair of evaluators. According to the interpretation guidelines from \cite{Landis1977TheMO}, the Cohen’s Kappa coefficients for Naturalness and Relevance (ranging from 0.79 to 0.85) fall within the \enquote{Substantial} to \enquote{Almost Perfect} categories, indicating strong inter-rater reliability for these metrics. This indicates that the evaluators share a common understanding of the evaluation criteria, leading to consistent ratings across evaluators. For Specificity, Novelty, and Actionability, the coefficients range from 0.58 to 0.68, placing them primarily in the \enquote{Moderate} to \enquote{Substantial} categories. These results suggest moderate reliability for these metrics, likely due to subjective interpretation and less clearly defined evaluation guidelines. Novelty, with lower coefficients around 0.61 to 0.63, highlights variability in ratings, suggesting that evaluators may have differing perspectives on what qualifies as novel (but the inter-rater reliability is still considered \enquote{Substantial}). Meanwhile, Actionability performs slightly better, nearing the \enquote{Substantial} range, indicating moderately consistent criteria.

\begin{table}[ht]
\centering
\resizebox{\linewidth}{!}{
\begin{tabular}{lccc}
\toprule
\textbf{Metric}      & \textbf{Evaluator 1 \& 2} & \textbf{Evaluator 1 \& 3} & \textbf{Evaluator 2 \& 3} \\ 
\midrule
Naturalness          & 0.85                     & 0.83                     & 0.84                     \\
Relevance            & 0.81                     & 0.79                     & 0.80                     \\
Specificity          & 0.65                     & 0.63                     & 0.66                     \\
Novelty              & 0.60                     & 0.58                     & 0.61                     \\
Actionability        & 0.67                     & 0.65                     & 0.68                     \\ 
\bottomrule
\end{tabular}}
\caption{Cohen's Kappa coefficient for various metrics.}
\label{tab:evaluator_agreement}
\end{table}

\subsection{Personalized PageRank (PPR)}
\label{sec:ppr}

In this section, we discuss the details of the PPR algorithm used in \refsec{sec:sub_graph_retrieval} to prune the graph from DBPedia and concentrate on nodes most pertinent to the central nodes of interest. The PPR is calculated using the iterative formula:
\begin{equation}
    \mathbf{p} = \alpha \mathbf{A}^\top \mathbf{p} + (1 - \alpha) \mathbf{s},
\end{equation}
where $\mathbf{p} \in \mathbb{R}^n$ is the PPR vector representing the relevance scores of $n$ nodes in the graph. $\alpha$ is the damping factor controlling the probability of continuing the random walk versus restarting from the personalization vector. $\mathbf{A}^\top$ is the transpose of the column-normalized adjacency matrix $\mathbf{A}$ of the graph, representing transition probabilities between nodes. $\mathbf{s} \in \mathbb{R}^n$ is the personalization vector, where we assign a value of 1 to the central nodes and 0 to all other nodes to emphasize their importance. To ensure convergence and computational efficiency, we set a tolerance parameter $\text{tol} = 1 \times 10^{-6}$ and a maximum iteration limit $\text{max\_iter} = 100$. After computing the PPR vector $\mathbf{p}$, we apply a threshold of $1 \times 10^{-5}$ to prune the graph. Nodes with PPR scores below this threshold are considered insignificant with respect to the central nodes and are thus removed. This process effectively filters out less relevant nodes, resulting in a pruned graph that highlights the most significant relationships and structures pertinent to our analysis.

\begin{table*}[ht]
    \centering
    \resizebox{\linewidth}{!}{
    \begin{tabular}{lcccccc}
        \toprule
        \textbf{Setting} & \textbf{Context Relevance} & \textbf{Comprehensiveness} & \textbf{Correctness} & \textbf{Empowerment} & \textbf{SAFE} & \textbf{FActScore} \\
        \midrule
        OKGQA (subgraphs) & $75.70\%\pm0.44\%$ & $71.51\%\pm0.83\%$ & $66.43\%\pm0.76\%$ & $69.60\%\pm0.65\%$ & $88.71\%\pm0.72\%$ & $70.12\%\pm0.87\%$ \\
        + Multi-lingual context & $75.14\%\pm0.33\%$ & $72.32\%\pm0.19\%$ & $66.72\%\pm0.74\%$ & $70.32\%\pm0.57\%$ & $90.32\%\pm0.48\%$ & $72.83\%\pm0.93\%$ \\
        \bottomrule
    \end{tabular}}
    \caption{\small Comparison of GPT-4o-mini Performance Using Monolingual and Multilingual Subgraphs}
    \label{tab:multilingual_comparison}
\end{table*}

\subsection{Prize-Cost-based Path Retrieval}
\label{sec:path-retrieval}

In this section, we detail the path-retrieval method used in \refsec{sec:g-retrieval}. It is designed to construct and evaluate paths in a graph based on predefined prize assignments and cost allocations. The objective is to form sequences of nodes and edges, represented as \( \mathcal{P} = \{v_1, e_1, v_2, \dots, e_{n-1}, v_n\} \), that maximize the overall score and minimize the costs.
To efficiently manage the exploration of potential paths, we utilize a \textbf{priority queue}, a data structure that allows paths to be organized based on their scores, ensuring that the highest-scoring paths are processed first. 
The method starts by picking a number of starting nodes with high prizes.From each starting node, it expands by exploring neighboring nodes. For each neighbor, a new score is computed as the sum of the neighbor’s prize and the edge’s prize minus the edge’s cost. If this neighbor has not been visited before, the algorithm appends it to the current path and adds this extended path to the priority queue. This expansion process continues until paths reach a maximum length or can no longer be extended. The algorithm maintains a record of explored paths to avoid repetition and cycles. Once no further expansions are possible or the priority queue is empty, the algorithm sorts the collected paths in descending order of their scores.

\subsection{LLM Evaluation Clarity}
\label{sec:llm_evaluation_clarity}

To address the concern regarding potential self-enhancement bias in LLM-as-evaluator frameworks, we provide extensive validation of our evaluation approach. In specific, we randomly sample 100 questions and evaluated them using three different LLMs (i.e., gpt-4o-mini, llama-3.1-8b-instruct, and gemma-2-9b-it). 
We measure the inter-model agreement using Cohen’s Kappa as shown in Table~\ref{tab:llm_metrics}. It indicates that the evaluation results are consistent across different LLMs, even when the model generating the responses is not the same as the one evaluating them (e.g., using gpt-4o-mini for generation and llama-3.1-8b-instruct for evaluation). These findings confirm that the evaluation is robust and independent of the specific LLM used as the evaluator. 
In addition, we also collect human evaluations for these 100 samples. Three expert annotators rate each anonymized response on context relevance, comprehensiveness, correctness, and empowerment using a 1–5 Likert scale. The average human ratings are then computed and compared with automated scores obtained via G-Eval. The Pearson correlation coefficient of 0.78 indicates strong alignment between human judgments and LLM-based evaluations. Together with the inter-model agreement reported in Table~\ref{tab:llm_metrics}, these results demonstrate that our evaluation is robust, consistent, and largely independent of the specific LLM used as the evaluator.

\begin{table}[ht]
    \centering
    \resizebox{\linewidth}{!}{
    \begin{tabular}{lccc}
    \toprule
    Metric      & LLM 1 \& 2 & LLM 1 \& 3 & LLM 2 \& 3 \\
    \midrule
    G-Eval      & 0.84       & 0.81       & 0.82       \\
    FactScore   & 0.78       & 0.74       & 0.78       \\
    SAFE        & 0.74       & 0.70       & 0.72       \\
    \bottomrule
    \end{tabular}}
    \caption{\small Cohen’s Kappa coefficient for different LLM pair comparisons. For the G-Eval, we use the average score of four sub-metrics for better readability. LLM 1: gpt-4o-mini; LLM 2: llama-3.1-8b-instruct; LLM 3: gemma-2-9b-it)}
    \label{tab:llm_metrics}
\end{table}

\subsection{KG Similarity Metrics}
\label{appendix:kg-similarity-metric}

In this section, we introduce the metrics used in \refsec{sec:method} to measure the deviation of the perturbed KGs from the original KG. These metrics are adapted from ~\cite{raman2020learning} as presented below. ATS is mainly used to measure the semantic similarity between two KGs, while SC2D and SD2 are used to measure the structural similarity.

\paragraph{Aggregated Triple Score (ATS):}
ATS measures semantic similarity between two KGs. Let \begin{small}$s_\mathcal{G}$\end{small} be an edge (triple) scoring function, such that \begin{small}$s_{\mathcal{G}}(e_1, r, e_2)$\end{small} measures how likely edge \begin{small}$(e_1, r, e_2)$\end{small} is to exist in \begin{small}$\mathcal{G}$\end{small}. Also, assume \begin{small}$s_\mathcal{G}$\end{small} has been pre-trained on \begin{small}$\mathcal{G}$\end{small} for link prediction. Then, ATS is defined as \begin{small}$f_{\text{ATS}}(\mathcal{G},\mathcal{G}^{'}) = \frac{1}{|\mathcal{T}^{'}|}\sum_{(e_1,r,e_2)\in\mathcal{T}^{'}}s_\mathcal{G}(e_1,r,e_2) \in [0, 1]$\end{small}, which denotes the mean \begin{small}$s_\mathcal{G}$\end{small} score across all edges in \begin{small}$\mathcal{G'}$\end{small}. Intuitively, if a high percentage of edges in \begin{small}$\mathcal{G'}$\end{small} are also likely to exist in \begin{small}$\mathcal{G}$\end{small} (i.e., high ATS), then we say that \begin{small}$\mathcal{G'}$\end{small} and \begin{small}$\mathcal{G}$\end{small} have high semantic similarity. $s_{\mathcal{G}}$ is task-specific, as KGs from different tasks may differ greatly in semantics. We use the $s_{\mathcal{G}}$ from \citep{li-etal-2016-commonsense}; while ATS captures semantic KG differences, it is not sensitive to KG connectivity structure. Note that \begin{small}$f_{\text{ATS}}(\mathcal{G},\mathcal{G})$\end{small} may not equal 1, since \begin{small}$s_\mathcal{G}$\end{small} may not perfectly generalize to KGs beyond those it was trained on.
% For instance, out of our five proposed perturbation methods, we can intuitively tell that ED destroys the most KG information, yet ED's ATS scores are consistently the best. 

\paragraph{Similarity in Clustering Coefficient Distribution (SC2D):}
SC2D measures structural similarity between two KGs and is derived from the local clustering coefficient \citep{saramaki2007generalizations, onnela2005intensity, fagiolo2007clustering}. For a given entity in \begin{small}$\mathcal{G}$\end{small} (treated here as undirected), the local clustering coefficient is the fraction of possible triangles through the entity that exist (i.e., how tightly the entity's neighbors cluster around it). For entity \begin{small}$e_{i} \in \mathcal{E}$\end{small}, the local clustering coefficient is defined as \begin{small}$c_{i} = 2 \text{Tri}(e_{i}) / (\text{deg}(e_{i})(\text{deg}(e_{i}) - 1)) $\end{small}, where \begin{small}$\text{Tri}(e_{i})$\end{small} is the number of triangles through \begin{small}$e_{i}$\end{small}, and \begin{small}$\text{deg}(e_{i})$\end{small} is the degree of \begin{small}$e_{i}$\end{small}. For each relation \begin{small}$r \in \mathcal{R}$\end{small}, let \begin{small}$\mathcal{G}^r$\end{small} be the subgraph of \begin{small}$\mathcal{G}$\end{small} consisting of all edges in \begin{small}$\mathcal{T}$\end{small} with \begin{small}$r$ \end{small}. That is, \begin{small}$\mathcal{G}^r = (\mathcal{E}, r, \mathcal{T}^{'})$, where $\mathcal{T}^{'}=\{(e, r, e') \hspace{1mm} | \hspace{1mm} e, e' \hspace{-0.5mm} \in \hspace{-0.5mm} \mathcal{E}\}$\end{small}. Let \begin{small}$\mathbf{c}^r$\end{small} denote the \begin{small}$|\mathcal{E}|$\end{small}-dimensional clustering coefficient vector for \begin{small}$\mathcal{G}^r$\end{small}, where the \begin{small}$i$\end{small}th element of \begin{small}$\mathbf{c}^r$\end{small} is \begin{small}$c_{i}$\end{small}. Then, the mean clustering coefficient vectors for \begin{small}$\mathcal{G}$\end{small} and \begin{small}$\mathcal{G}'$\end{small} are \begin{small}$\mathbf{c}_o = \frac{1}{|\mathcal{R}|}\sum_{r \in \mathcal{R}}\mathbf{c}^r$\end{small} and \begin{small}$\mathbf{c}_p = \frac{1}{|\mathcal{R}'|}\sum_{r \in \mathcal{R}'}\mathbf{c}^r$\end{small}, respectively. SC2D is defined as \begin{small}$f_{\text{SC2D}}(\mathcal{G},\mathcal{G}^{'}) = 1 - \frac{\lVert \mathbf{c}_{o} - \mathbf{c}_{p} \rVert_{2}}{\lVert \mathbf{c}_{o} - \mathbf{c}_{p} \rVert_{2} + 1} \in [0, 1]$\end{small}, with higher value indicating higher similarity.
% Finally, SC2D is defined as \begin{small}$f_{\text{SC2D}}(\mathcal{G},\mathcal{G}^{'}) = \frac{1}{\lVert \mathbf{c}_{o} - \mathbf{c}_{p} \rVert_{2} + {\epsilon}}$\end{small}, where \begin{small}$\epsilon$\end{small} is a small constant to avoid division by zero.

\paragraph{Similarity in Degree Distribution (SD2):}
SD2 also measures structural similarity between two KGs, while addressing SC2D's ineffectiveness when the KGs' entities have tiny local clustering coefficients (e.g., the item KG used by recommender systems is roughly bipartite). In such cases, SC2D is always close to one regardless of the perturbation method, thus rendering SC2D useless. Let  \begin{small}$\mathbf{d}^{r}$\end{small} denote the \begin{small}$|\mathcal{E}|$\end{small}-dimensional degree vector for \begin{small}$\mathcal{G}^r$\end{small}, where the \begin{small}$i$\end{small}th element of \begin{small}$\mathbf{d}^r$\end{small} is \begin{small}$\text{deg}(e_i)$\end{small}. Then, the mean degree vectors for \begin{small}$\mathcal{G}$\end{small} and \begin{small}$\mathcal{G}'$\end{small} are \begin{small}$\mathbf{d}_o = \frac{1}{|\mathcal{R}|}\sum_{r \in \mathcal{R}}\mathbf{d}^r$\end{small} and \begin{small}$\mathbf{d}_{p} = \frac{1}{|\mathcal{R}'|}\sum_{r \in \mathcal{R}'}\mathbf{d}^r$\end{small}, respectively. SD2 is defined as \begin{small}$f_{\text{SD2}}(\mathcal{G},\mathcal{G}^{'}) = 1 - \frac{\lVert \mathbf{d}_{o} - \mathbf{d}_{p} \rVert_{2}}{\lVert \mathbf{d}_{o} - \mathbf{d}_{p} \rVert_{2} + 1} \in [0, 1]$\end{small}, with higher value indicating higher similarity.

\section{Extension of \ours{}}
\label{sec:extension_of_benchmark}

In this section, we extend our benchmark by incorporating multilingual context and validating our query generation against DBpedia’s structure. We first introduce the multilingual setup of our dataset anc compare the performance of multilingual subgraphs with the monolingual subgraphs (\refsec{sec:multilingual_setup}). We then analyze the relationship between generated queries and DBpedia by examining query generation, entity/relation coverage, and subgraph alignment (\refsec{sec:relationship_queries_dbpedia}). We also compare \ours{} with the existing widely used benchmarks in Table~\ref{tab:benchmark_comparison}.

\subsection{Multilingual Setup of \ours{}}
\label{sec:multilingual_setup}
KGs typically include entities and relations in multiple languages, providing a richer context that can benefit our OKGQA setting. In this experiment, we investigate whether incorporating multilingual context improves overall performance. Specifically, we randomly sample 300 queries from our dataset and generate subgraphs that include multilingual entities and relations from DBpedia. We then apply PPR consistent with our original method in \refsec{sec:dataset_construction} to reduce the KG size. For this multilingual setting, we consider five languages—Greek, Polish, Portuguese, Spanish, and English—which cover the majority of entities in DBpedia.
We compare the performance of GPT-4o-mini using the new multilingual subgraphs against the original monolingual subgraphs, as shown in Table~\ref{tab:multilingual_comparison}. Our findings indicate that including multilingual context generally leads to better performance across multiple metrics. Intuitively, this additional multi-lingual context may provides more knowledge from different perspectives (which could provide more context, but also may requires more techniques for handle challenges like duplicates across languages) and also provide another way to validate the factuality of the resources stored in the KGs (which can provide more authenticity through cross validation from different languages).

\subsection{Generated Query-DBpedia Alignment}
\label{sec:relationship_queries_dbpedia}

We analyze the alignment between our generated queries and DBpedia along three dimensions: query generation, entity/relation coverage, and subgraph alignment as follows:

\paragraph{Query Generation:}  
Each query is directly generated from DBpedia entities and their relationships. For example, when asking about Microsoft's founder, we first confirm that both \enquote{Microsoft} and \enquote{Bill Gates} exist in DBpedia and are connected by the \texttt{founded\_by} relation, ensuring that our queries are firmly grounded in the knowledge graph.

\paragraph{Entity and Relation Coverage:}
Our analysis indicates that: 
\begin{itemize} 
\setlength\itemsep{0em}
\item 92\% entities mentioned in the queries can be detected from DBpedia entities. 
\item 87\% queries have complete relation paths connecting the relevant entities from DBPedia. 
\item Entities/relations mentioned in queries cover 72\% of DBpedia's most common entities/predicates and span diverse entity types (e.g., Person, Organization, and Event). 
\end{itemize}

\paragraph{Subgraph Alignment:}  
We evaluate the structure of the sampled subgraphs for each query and find that:
\begin{itemize}
\setlength\itemsep{0em}
\item 75\% of the queries retrieve subgraphs within 3–4 hops, which aligns with the typical depth for DBpedia reasoning tasks.
\item On average, each subgraph contains 48 nodes and 152 edges, with an average node degree of 3.17 and a clustering coefficient of 0.69, which also aligns with the property of DBPedia.
\end{itemize}

These statistics support that our dataset accurately reflects DBpedia’s structure, ensuring both authenticity and complexity in the generated queries.

\begin{table*}[t]
\centering
\resizebox{\textwidth}{!}{
\begin{tabular}{l c l p{9cm} l l c c}
\toprule
\textbf{Dataset} & \textbf{\# Questions} & \textbf{Question Type} & \textbf{Focus Areas} & \textbf{Source of Questions} & \textbf{Knowledge Base} & \textbf{ Hallucination Detection} & \textbf{Unreliable KG} \\
\midrule
OKGQA              & 850 / 2,050         & Open-ended     & Evaluating hallucination and reasoning capabilities in LLMs when augmented with Knowledge Graphs; diverse queries requiring complex reasoning & Curated & DBPedia & \cmark{} & \cmark \\
            [2pt]\hdashline\\[-8pt]

WebQuestions       & 5,810               & Factoid        & Questions derived from Google Suggest queries, focusing on simple factual information & User queries & Freebase & \xmark & \xmark \\
            [2pt]\hdashline\\[-8pt]

ComplexWebQuestions& 34,689              & Multi-hop Factoid & Extends WebQuestions with more complex, multi-hop questions requiring compositional reasoning & User queries & Freebase & \xmark & \xmark \\
            [2pt]\hdashline\\[-8pt]

GrailQA            & 64,331              & Varied Factoid & Evaluates generalization in KBQA with questions requiring different levels of reasoning & Crowdsourced & Freebase & \xmark & \xmark \\
\bottomrule
\end{tabular}}
\caption{\small Comparison of \ours{} with existing benchmarks along with their question types, focus areas, and additional properties.}
\label{tab:benchmark_comparison}
\end{table*}

\section{Related Work}
\label{sec:related_work}

% \textbf{Existing benchmarks in KGQA.} 
% \textbf{Integration of Knowledge Graphs into LLMs.}

% Integrating external knowledge sources, such as Knowledge Graphs (KGs) has been an emerging research trend. We classify the strategies for enhancing LLMs with KGs into three categories: \textit{Knowledge-aware} \textbf{Inference}, \textit{Knowledge-aware} \textbf{Learning} and \textit{Knowledge-aware} \textbf{Validation}. Each uniquely contributes to the refinement of a series of methods to improve trustworthiness/performance of the LLMs, including the inference process, improving the learning mechanism, and establishing robust methods for validating the model's decisions.
% % \subsection{Knowledge-aware Inference}
% In LLMs, \textit{``inference"} means generating text or predictions from a pre-trained model base on an input context. Challenges include incorrect or sub-optimal outputs due to ambiguous inputs, unclear context, knowledge gaps, training data biases, or inability to generalize to unseen scenarios. In addition, LLMs are often struggle with multi-step reasoning and, un-like humans, can not seek extra information to clarify ambiguous queries. To improve LLMs' inferences and reasoning, researchers integrate KGs at the input level to enhance contextual understanding.

Due to the stochastic decoding process of Large Language Models (LLMs), \textit{i.e.}, sampling the next token in the sequence, LLMs exhibit probabilistic behaviors: (1) potentially yielding varied outputs of the same input across different instances~\citep{agrawal2023hallucinationsurvey}; (2) cannot accurately interpret phrases or terms when the context is vague and resides in a knowledge gap of the model. This will lead to outputs that may sound plausible but are often irrelevant or incorrect. This will lead to outputs that may sound plausible but are often irrelevant or incorrect. This \enquote{hallucinations} undermines the reliability of LLMs~\citep{huang2023survey}.
One emerging research trend is enhancing LLMs through integrating external knowledge graphs~\citep{agrawal2023hallucinationsurvey}. 
KGs offer structured, explicit, and up-to-date factual knowledge, including domain-specific knowledge, providing a faithful knowledge source for reasoning~\citep{sui2021question, zheng2023does, agrawal2023hallucinationsurvey}. Moreover, each piece of information in KGs can be traced back to its source, providing context and provenance. This traceability not only aids in verifying the reliability of the information but also provides clear pathways of reasoning.

Researchers employ diverse strategies to augment the LLMs by integrating external KGs~\citep{sui2024fidelisfaithfulreasoninglarge,he2024unigraph,he2024unigraph2,SUI2022108943}. For example, 
KAPING~\citep{baeketal2023knowledge} matches entities in questions to retrieve related triples from knowledge graphs for zero-shot question answering. \citet{wu2023retrieve} finds that converting these triples into textualized statements can further enhance LLM performance. StructGPT~\citep{jiang2023structgpt} propose to convert user query into structured formats (e.g., SPARQL) for information extraction from KGs. Following the succuess of internal reasoning-enhancement methods like Chain-of-thoughts (CoT)~\citep{wei2022cot}, Reflexion~\citep{shinn2024reflexion}, and Tree-of-thoughts (ToT), \citet{he2022rethinking} propose \enquote{rethinking with retrieval} to use decomposed reasoning steps from CoT prompting to retrieve external knowledge, leading to more accurate and faithful explanations. IR-CoT~\citep{trivedi2022ircot} interleaves the generation of CoT with knowledge retrieval from corresponding KGs, iteratively guiding both retrieval and reasoning for multi-step questions. MindMap~\citep{wen2023mindmap} introduce a plug-and-play approach to evoke graph-of-thoughts reasoning in LLMs. Similarly, RoG~\citep{luo2023rog} use KGs to create faithful reasoning paths based on various relations, enabling interpretable reasoning.

However, current benchmarks for testing the capabilities of these LLM+KG models are predominantly closed-ended, restricting responses to a limited set of entities/relations or a set of logical forms derived from specific facts of KG. Hence, they can only test a very limited subset of the LLM's tendency to hallucinate, leaving a gap in the assessment of complex, real-world scenarios.
Particularly, standard metrics such as FActScore~\citep{minetal2023factscore} and SAFE~\citep{wei2024longformfactualitylargelanguage} for evaluating the hallucination rate of LLMs require open-ended settings, \textit{i.e.}, questions are phrased as a statement which requires a longer answer. Compared with previous works, our proposed \ours{} is tailored for evaluating LLMs enhanced with KGs under open-ended, real-world question-answering scenarios. The benchmark extends the assessment of closed-ended question answering to an open-ended setting, which can further support the assessment of hallucination of LLMs.

% Recent advancements in retrieval-augmented generation (RAG) have introduced innovative strategies to mitigate knowledge hallucination in large language models (LLMs). Coarse-to-Fine Highlighting~\cite{lv2024coarsetofinehighlightingreducingknowledge} utilizes a hierarchical filtering process to progressively refine retrieved knowledge, ensuring that only the most relevant and accurate information is used, making it particularly effective for addressing hallucinations in complex queries. SuRe~\cite{kim2024suresummarizingretrievalsusing} focuses on summarizing retrieved documents into concise, query-relevant representations, enhancing both computational efficiency and the contextual alignment of knowledge with the query. Similarly, RECOMP~\cite{xu2023recompimprovingretrievalaugmentedlms} introduces dynamic context compression and selective augmentation, tailoring retrieved information to the query’s specific needs to improve response quality and relevance. Collectively, these methods underscore the importance of balancing retrieval breadth with precision and computational efficiency in RAG systems. Their principles align closely with the goals of integrating knowledge graphs (KGs) to reduce hallucination in LLMs, as explored in OKGQA. By adopting techniques such as hierarchical filtering, summarized retrieval, and adaptive augmentation, future KG-augmented frameworks can enhance the reliability and accuracy of open-ended QA tasks, even when dealing with noisy or incomplete knowledge.

\section{Prompt List}
\label{sec:prompt_list}

In this section, we present all the prompts required for the main experiments. To enhance clarity, we provide only one example in the prompt labeled as \textcolor{b}{\texttt{Example 1}}; the other few-shot examples utilized are labeled as \textcolor{b}{\texttt{Other In-Context Few-shots}} within the prompt.

\subsection{Knowledge-Augmented Generation}
\label{prompt:f1}

\textbf{System Instruction:}
\enquote{You are a helpful assistant designed to answer the users' open-ended questions. Your task is to provide accurate, concise, and useful information to foster understanding and solve problems. Whether the questions relate to complex scientific concepts, historical events, practical advice, or everyday life, your goal is to assist by offering thoughtful and informative responses.}

\noindent{\textcolor{b}{\texttt{In-Context Few-shots}}}

\noindent\textbf{Question}: \textcolor{r}{\{Question\}}

\noindent\textbf{Below are the facts that might be relevant to answer the question:} \textcolor{g}{\{Retrieved-knowledge\}}

\noindent\textbf{Answer}:

% \section{Extra Hallucination Metrics} To quantify model hallucinations, \citep{lee2022factualityprompt} proposed two evaluation metrics:
% \begin{itemize}
%     \item \textbf{Hallucination NE (Named Entity) errors}: Using a pretrained entity detection model and document-level grounding, this metric measures the fraction of detected named entities that do not appear in the ground truth document.
%     \item \textbf{Entailment ratios}: Using a RoBERTa model fine-tuned on MNLI and sentence-level knowledge grounding, this metric calculates the fraction of generated sentences that are marked as relevant to the paired Wikipedia sentence by the entailment model.
% \end{itemize}
% High NE errors and low entailment ratios indicate higher factuality, and both metrics are found to be correlated with human annotations. Larger models are found to perform better on this benchmark.

\subsection{\ours{} Query Generation Prompting}
\label{prompt:f2}

\textbf{System Instruction:}
\enquote{Generate open-ended questions about different types: character description, event description, cause explanation, relationship explanation, trend prediction, outcome prediction, contrast analysis, historical comparison, evaluation and reflection, and application and practice. Please provide specific suggestions. Generate the questions, the type of the questions, the placeholders, the naturalness of your generated questions (choose from high, medium, and unnatural), the difficulty of the generated questions (choose from hard, medium and easy) and DBPedia\_entities (link the placeholders to DBPedia entities) in JSON format.}

\begin{figure}
    \centering
    \includegraphics[width=0.95\linewidth]{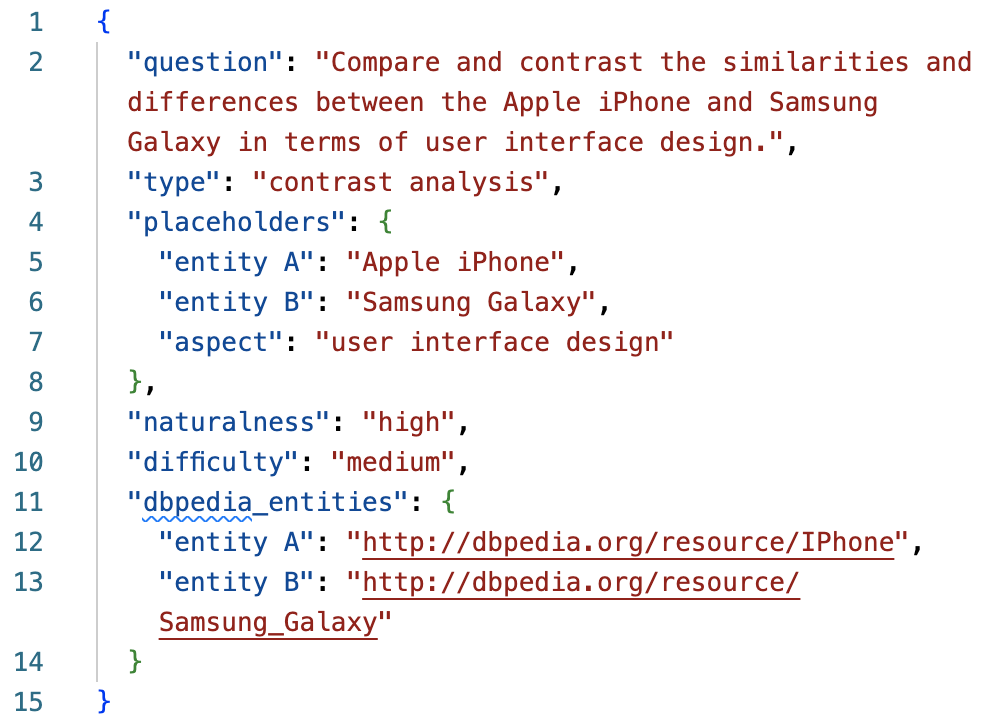}
    \caption{Example 1 Demonstration.}
    \label{listing_json}
\end{figure}

\noindent\textcolor{b}{\texttt{Example 1}}: as shown in Figure~\ref{listing_json}.

\noindent\textcolor{b}{\texttt{Other In-Context Few-shots}}

\noindent\textbf{Generation}:

\subsection{Prompts for Instruction Tuner}
\label{prompt:instruction_tuner}
Act as an \enquote{Instruction Tuner} for the LLM, you will be given the inputs: (1) the \textcolor{r}{\{Current Instruction\}} used to guide the LLMs's evaluation, including specific examples with ground truth labels; (2) \textcolor{r}{\{Current Errors\}} that emerged with this instruction are applied to the dataset.

The current errors are presented in the following format: (1) INPUT: \textcolor{r}{\{input text\}} (2) PREDICTED OUTPUT: \textcolor{r}{\{predicted label\}}, (3) EXPECTED OUTPUT: \textcolor{r}{\{ground truth label\}}. Carefully analyze these errors and craft a revised concise instruction for the LLM to fit the expected outputs. Include 2-3 examples at the end of your response to demonstrate how the new instruction would be applied.

\subsection{Metrics Prompt for G-eval}

\textbf{System Instruction:} \enquote{You are a helpful assistant designed to evaluate the quality of the response to a query. Your task is to rate the response on one metric defined as below:}

\noindent\textcolor{b}{\texttt{Empowerment Criteria}}:
Evaluate whether the \enquote{Actual Output} can help the reader understand the topic and make informed decisions regarding the \enquote{Input}. A response with high empowerment provides accurate information and explanations that enhance the reader’s understanding. When evaluating empowerment, consider the relevance of the information provided in the \enquote{Actual Output} to the \enquote{Input} and the \enquote{Retrieval Context}.

\noindent\textcolor{b}{\texttt{Comprehensiveness Criteria}}: Evaluate the extent to which the \enquote{Actual Output} covers all aspects and details of the question \enquote{Input}. A comprehensive answer should thoroughly address every part of the question, leaving no important points unaddressed. When evaluating comprehensiveness, consider the relevance of the information provided in the \enquote{Actual Output} to the \enquote{Input} and the \enquote{Retrieval Context}.

\noindent\textcolor{b}{\texttt{Correctness Criteria}}: Measure how clearly and specifically the \enquote{Actual output} responds to the question \enquote{input}. A highly direct response stays focused on the question, providing clear and unambiguous information. When evaluating correctness, consider the relevance of the information provided in the \enquote{Actual Output} to the \enquote{Input} and the \enquote{Retrieval Context}.

\noindent\textcolor{b}{\texttt{Context Relevance Criteria}}: Evaluate the extent to which the \enquote{Actual output} incorporates relevant information from the \enquote{Retrieval Context}. This includes assessing whether the output adheres to the thematic, factual, and situational specifics presented in the \enquote{Retrieval Context}. Relevant responses not only address the direct query but also align closely with the contextual elements provided, ensuring a seamless and coherent transition between the \enquote{Retrieval Context} and the \enquote{Actual Output}. The most contextually relevant responses demonstrate an understanding and appropriate reflection of the given circumstances, historical facts, or conceptual background, thereby contributing to the overall accuracy and utility of the information provided.

\noindent\textbf{Response}: [Respond with metric and the corresponding score.]

\end{document}